\documentclass[10pt,twocolumn,letterpaper]{article}

\usepackage{iccv}
\usepackage{times}

\iccvfinalcopy 

\usepackage{graphicx}
\usepackage{amsmath}
\usepackage{amssymb}
\usepackage{booktabs}
\usepackage{subfig}
\usepackage[font=small]{caption}

\usepackage{diagbox}
\usepackage{nicefrac}       %
\usepackage{microtype}      %

\usepackage{blindtext}
\usepackage{float}
\usepackage{enumitem}
\usepackage[table]{xcolor}
\usepackage{tabulary,multirow,overpic}
\usepackage{verbatim}
\usepackage{color}
\usepackage{multicol}
\usepackage{dblfloatfix}
\usepackage{pifont}%
\usepackage[accsupp]{axessibility}  %
\usepackage{makecell}

\usepackage{multicol}

\usepackage{fontawesome5}

\newlength\savewidth\newcommand\shline{\noalign{\global\savewidth\arrayrulewidth\global\arrayrulewidth 1pt}\hline\noalign{\global\arrayrulewidth\savewidth}}
\newcommand{\tablestyle}[2]{\setlength{\tabcolsep}{#1}\renewcommand{\arraystretch}{#2}\centering\footnotesize}

\newcolumntype{x}[1]{>{\centering\arraybackslash}p{#1pt}}
\newcolumntype{y}[1]{>{\raggedright\arraybackslash}p{#1pt}}
\newcolumntype{z}[1]{>{\raggedleft\arraybackslash}p{#1pt}}

\definecolor{baselinecolor}{gray}{.92}

\definecolor{demphcolor}{gray}{.2}

\definecolor{demphcolorinline}{gray}{.3}

\definecolor{demphcolor1}{gray}{.6}

\definecolor{eva01purple}{RGB}{168,119,200}
\newcommand{\evapurple}[1]{\textcolor{eva01purple}{#1}}

\definecolor{eva01green}{RGB}{82,208,83}
\newcommand{\evagreen}[1]{\textcolor{eva01green}{#1}}

\definecolor{eva02red}{RGB}{236,35,35}
\newcommand{\evared}[1]{\textcolor{eva02red}{#1}}

\definecolor{eva02yellow}{RGB}{249,157,83}

\definecolor{02pink}{RGB}{240,178,188}

\definecolor{00blue}{RGB}{100,149,237}
\newcommand{\evablue}[1]{\textcolor{00blue!80}{#1}}

\newcommand{\ph}[1]{\textcolor{white}{#1}}

\newcommand{\phgray}[1]{\textcolor{Graylight!30}{#1}}

\definecolor{citecolor}{RGB}{34,139,34}
\definecolor{citecolor2}{HTML}{0071bc}
\definecolor{Graylight}{gray}{0.9}
\definecolor{lightred}{RGB}{241,140,142}
\usepackage[pagebackref=true,breaklinks=true,colorlinks,
citecolor=eva02red,bookmarks=false]{hyperref}

\definecolor{clipbaselinecolor}{gray}{.9}

\definecolor{defaultcolor}{HTML}{E8E2F7}

\renewcommand{\paragraph}[1]{\vspace{1.25mm}\noindent\textbf{#1}}

\newcommand{\app}{\raise.17ex\hbox{$\scriptstyle\sim$}}
\newcommand{\appp}{\raise.20ex\hbox{$\scriptscriptstyle\sim$}}

\def\x{$\times$}

\newcommand{\tblref}[1]{Table~\ref{#1}}

\newcommand{\evaone}{{\textbf{\evapurple{EVA}}}\xspace}

\newcommand{\evaclip}{{\textbf{\evablue{EVA-CLIP}}}\xspace}
\newcommand{\evaOneclip}{{\textbf{\evablue{EVA-01-CLIP}}}\xspace}
\newcommand{\evaTwoclip}{{\textbf{\evablue{EVA-02-CLIP}}}\xspace}

\newcommand{\evaclipeight}{{\textbf{\evablue{EVA-CLIP-8B}}}\xspace}
\newcommand{\evaclipx}{{\textbf{\evablue{EVA-CLIP-18B}}}\xspace}

\newcommand{\rgray}{\rowcolor{Graylight!30}}

\newcommand{\suptext}[1]{$^{\text{#1}}$}

\usepackage[capitalize]{cleveref}
\crefname{section}{Sec.}{Secs.}
\Crefname{section}{Section}{Sections}
\Crefname{table}{Table}{Tables}
\crefname{table}{Tab.}{Tabs.}
\newcommand{\authorskip}{\hspace{4mm}}

\begin{document}

\title{\evaclipx: Scaling CLIP to 18 Billion Parameters}

\author{Quan Sun\textsuperscript{1}\thanks{Correspondence to \textit{wangxinlong@baai.ac.cn} } 
\authorskip Jinsheng Wang\textsuperscript{1}$^*$
\authorskip Qiying Yu\textsuperscript{1,2}$^*$
\authorskip Yufeng Cui\textsuperscript{1} \\
\authorskip Fan Zhang\textsuperscript{1} 
\authorskip Xiaosong Zhang\textsuperscript{1}
\authorskip Xinlong Wang\textsuperscript{1}
\\[2mm]
{
\fontsize{10.0pt}{9.84pt}\selectfont
\textsuperscript{1} Beijing Academy of Artificial Intelligence \hspace{5.5mm} \textsuperscript{2} Tsinghua University}\\[1mm]
{
\fontsize{9.4pt}{9.84pt}\selectfont  
\textsuperscript{$\ast$}equal contribution
} \\[2mm]
{
\fontsize{8.4pt}{9.84pt}\selectfont
code \& models: \href{https://github.com/baaivision/EVA/tree/master/EVA-CLIP-18B}{\color{00blue!80} \bfseries \ttfamily baaivision/EVA/EVA-CLIP-18B}
}
}

\maketitle

\begin{abstract}
    Scaling up contrastive language-image pretraining (CLIP) is critical for empowering both vision and multimodal models.
    We present \evaclipx, the largest and most powerful open-source CLIP model to date, with 18-billion parameters. With only 6-billion training samples seen, \evaclipx achieves an exceptional \textbf{80.7}\% zero-shot top-1 accuracy averaged across 27 widely recognized image classification benchmarks, outperforming its forerunner \evaclip (5-billion parameters) and other open-source CLIP models by a large margin. Remarkably, we observe a consistent performance improvement with the model size scaling of \evaclip, despite maintaining a constant training dataset of 2-billion image-text pairs from LAION-2B and COYO-700M.
    This dataset is openly available and much smaller than the in-house datasets (\textit{e.g.}, DFN-5B, WebLI-10B) employed in other state-of-the-art CLIP models.
    \evaclipx demonstrates the potential of EVA-style~\cite{eva,EVA02,sun2023evaclip} weak-to-strong visual model scaling. 
    With our model weights made publicly available, we hope to facilitate future research in vision and multimodal foundation models.

\end{abstract}

\section{Introduction}

Recent years witnessed the rapid growth of Large Multimodal Models (LMMs)~\cite{alayrac2022flamingo, emu1, emu2, wang2023cogvlm, Qwen-VL, liu2023llava}, with CLIP models ~\cite{clip, openclip2022scalelaws, sun2023evaclip, li2023clipav2, siglip, dfn5b, chen2023internvl} serving as a foundational vision encoder to deliver robust and transferable visual representations, and Large Language Models (LLMs)~\cite{touvron2023llama,t5} serving as a general interface for reasoning across different modalities. 
However, as LLMs have scaled up to around 100B parameters or higher~\cite{gpt3,chowdhery2022palm,touvron2023llama}, the adopted vision foundation models continue to operate at a much smaller scale, lagging far behind the LLMs. 

\begin{figure}[t]
    \centering
    \includegraphics[width=1.0\linewidth]{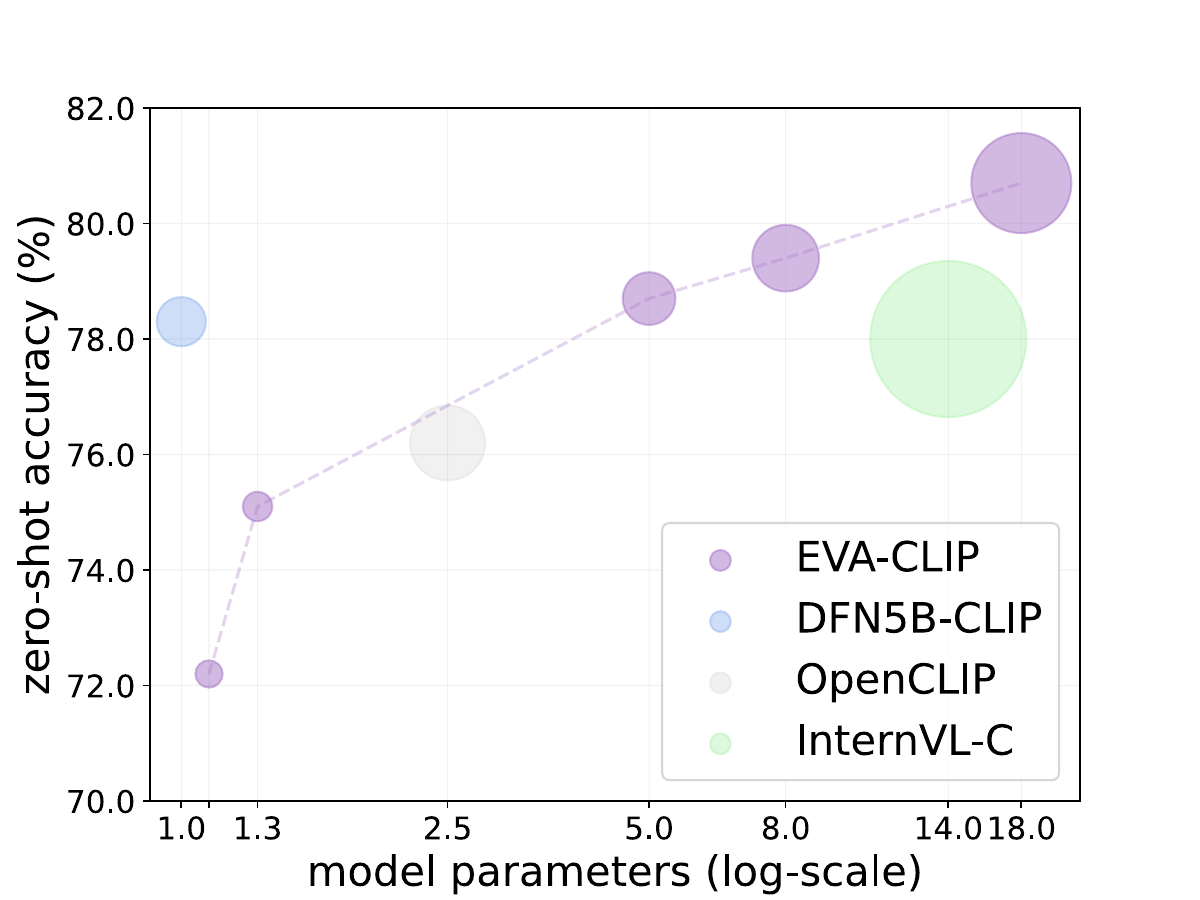}
    \caption{\textbf{Scaling behavior of \evaclip with zero-shot classification performance averaged across 27 image classification benchmarks}, compared with the current state-of-the-art and largest CLIP models (224px). The diameter of each circle demonstrates the forward GFLOPs $\times$ the number of training samples seen. The performance of \evaclip consistently improves as scaling up.}
    \label{fig:teaser}
\end{figure}

\begin{table*}[t!]
\vspace{-20pt}
\centering
\tablestyle{1.4pt}{1.2}
    \begin{tabular}{l|ccc|cc|ccc|c|c|c}
        & total & image & text & & samples & image & batch & & image cls. & video cls. & retrieval \\
        model{\scriptsize{\ph{+}}} & \#param. & \#param. & \#param. & data & seen & size & size & gpus for training & avg. acc. & avg. acc. & MR \\
        \shline

        \rgray
        \scriptsize {\evaOneclip-g/14\phgray{+}}~\cite{sun2023evaclip} & \scriptsize 1.1B & \scriptsize 1.0B & \scriptsize 124M & \scriptsize LAION-400M~\cite{laion400m} & \scriptsize 11B & \scriptsize 224\suptext{2} & \scriptsize 41k & \scriptsize 256{\scriptsize{\x}}A100 \scriptsize{(40GB)} & \scriptsize 72.2 & \scriptsize 66.2 & \scriptsize 80.9 \\
    
        \rgray
        \scriptsize {\evaOneclip-g/14+}~\cite{sun2023evaclip} & \scriptsize 1.3B & \scriptsize 1.0B & \scriptsize 354M & \scriptsize Merged-2B~\cite{sun2023evaclip} & \scriptsize 11B & \scriptsize 224\suptext{2} & \scriptsize 114k & \scriptsize 112{\scriptsize{\x}}A100 \scriptsize{(40GB)} & \scriptsize 75.1 & \scriptsize 68.8 & \scriptsize 85.3 \\
        
        \scriptsize OpenCLIP-G/14\ph{+}~\cite{clipbigg} & \scriptsize 2.5B & \scriptsize 1.8B & \scriptsize 695M & \scriptsize LAION-2B~\cite{laion5b} & \scriptsize 39B & \scriptsize 224\suptext{2} & \scriptsize 160k & \scriptsize 512{\scriptsize{\x}}A100 \scriptsize{(80GB)} & \scriptsize 76.2 & \scriptsize 68.7 & \scriptsize 85.7 \\
 
        \scriptsize InternVL-C\ph{+}~\cite{chen2023internvl} & \scriptsize 14.0B & \scriptsize 6.0B & \scriptsize 8.0B & \scriptsize custom~\cite{chen2023internvl} & \scriptsize 29B & \scriptsize 224\suptext{2} & \scriptsize 164k & \scriptsize 640{\scriptsize{\x}}A100 \scriptsize{(80GB)} & \scriptsize 78.0 & \scriptsize 73.7 & \scriptsize 86.6 \\

        \scriptsize DFN5B-CLIP-H/14\ph{+}~\cite{dfn5b} & \scriptsize 1.0B & \scriptsize 632M & \scriptsize 354M & \scriptsize DFN-5B~\cite{dfn5b} & \scriptsize 39B & \scriptsize 224\suptext{2} & \scriptsize 79k & \scriptsize TPUv4 & \scriptsize 78.3 & \scriptsize 67.0 & \scriptsize 86.6 \\
        
        \rgray
        \scriptsize {\evaTwoclip-E/14+}~\cite{sun2023evaclip} & \scriptsize 5.0B & \scriptsize 4.4B & \scriptsize 695M & \scriptsize LAION-2B~\cite{laion5b} & \scriptsize 9B & \scriptsize 224\suptext{2} & \scriptsize 144k & \scriptsize 144{\scriptsize{\x}}A100 \scriptsize{(80GB)} & \scriptsize 78.7 & \scriptsize 72.1 & \scriptsize 85.7 \\

        \scriptsize DFN5B-CLIP-H/14+~\cite{dfn5b} & \scriptsize 1.0B & \scriptsize 632M & \scriptsize 354M & \scriptsize DFN-5B~\cite{dfn5b} & \scriptsize 5B & \scriptsize 378\suptext{2} & \scriptsize 79k & \scriptsize TPUv4 & \scriptsize 79.2 & \scriptsize 68.4 & \scriptsize 87.2 \\

        \rgray
        \scriptsize {\evaclipeight\phgray{+}} & \scriptsize 8.1B & \scriptsize 7.5B & \scriptsize 695M & \scriptsize Merged-2B~\cite{sun2023evaclip} & \scriptsize 9B & \scriptsize 224\suptext{2} & \scriptsize 178k & \scriptsize 384{\scriptsize{\x}}A100 \scriptsize{(40GB)} & \scriptsize 79.4 & \scriptsize 73.6 & \scriptsize 86.2 \\

        \rgray
        \scriptsize {\evaclipx \phgray{+}} & \scriptsize 18.1B & \scriptsize 17.5B & \scriptsize 695M & \scriptsize Merged-2B+ & \scriptsize 6B & \scriptsize 224\suptext{2} & \scriptsize 108k & \scriptsize 360{\scriptsize{\x}}A100 \scriptsize{(40GB)} & \scriptsize  \textbf{80.7} & \scriptsize \textbf{75.0} & \scriptsize \textbf{87.8} \\
        
        \end{tabular}
\caption{\textbf{CLIP model configurations and zero-shot performance on 33 benchmarks including 27 image classification, 4 video classification and 2 image-text retrieval datasets.} DFN-5B~\cite{dfn5b} are 5B images filtered from a pool of 43B uncurated image-text pairs consisting of 12.8B image-text pairs from CommonPool-12.8B~\cite{gadre2023datacomp} and 30B additional public image-text pairs. The dataset used for training InternVL-C~\cite{chen2023internvl} is custom mixtures, see detail in \cite{chen2023internvl}.}
\label{tab: clip config and results }
\end{table*}

\begin{table*}[t!]
\vspace{-10pt}
\centering
\tablestyle{1.4pt}{1.2}
    \begin{tabular}{l|cccccc|ccccccccccccccccccccc|c}
        \scriptsize method &
        \rotatebox[origin=l]{90}{\scriptsize{ImageNet-1K~\cite{deng2009imagenet}}} &
        \rotatebox[origin=l]{90}{\scriptsize{ImageNet-V2~\cite{recht2019imagenetv2}}} &
        \rotatebox[origin=l]{90}{\scriptsize{ImageNet-Adv.~\cite{inadv}}} &
        \rotatebox[origin=l]{90}{\scriptsize{ImageNet-Ren.~\cite{inren}}} &
        \rotatebox[origin=l]{90}{\scriptsize{ImageNet-Ske.~\cite{inske}}} &
        \rotatebox[origin=l]{90}{\scriptsize{ObjectNet~\cite{objectnet}}} &
        \rotatebox[origin=l]{90}{\scriptsize{CIFAR-10~\cite{cifar}}} &
        \rotatebox[origin=l]{90}{\scriptsize{CIFAR-100~\cite{cifar}}} & 
        \rotatebox[origin=l]{90}{\scriptsize{MNIST~\cite{lecun1998gradient}}} & 
        \rotatebox[origin=l]{90}{\scriptsize{Caltech101~\cite{fei2004learning}}} & 
        \rotatebox[origin=l]{90}{\scriptsize{SUN397~\cite{xiao2010sun}}} & 
        \rotatebox[origin=l]{90}{\scriptsize{FGVC Aircraft~\cite{maji2013fine}}} & 
        \rotatebox[origin=l]{90}{\scriptsize{Country-211~\cite{clip}}} & 
        \rotatebox[origin=l]{90}{\scriptsize{Stanford Cars~\cite{krause20133d}}} &
        \rotatebox[origin=l]{90}{\scriptsize{Birdsnap~\cite{berg2014birdsnap}}} & 
        \rotatebox[origin=l]{90}{\scriptsize{DTD~\cite{cimpoi14describing}}} & 
        \rotatebox[origin=l]{90}{\scriptsize{EuroSAT~\cite{helber2019eurosat}}} & 
        \rotatebox[origin=l]{90}{\scriptsize{FER2013~\cite{goodfellow2013challenges}}} & 
        \rotatebox[origin=l]{90}{\scriptsize{Flowers-102~\cite{nilsback2008automated}}} & 
        \rotatebox[origin=l]{90}{\scriptsize{Food-101~\cite{bossard2014food}}} & 
        \rotatebox[origin=l]{90}{\scriptsize{GTSRB~\cite{stallkamp2012man}}} & 
        \rotatebox[origin=l]{90}{\scriptsize{PCam~\cite{veeling2018rotation}}} & 
        \rotatebox[origin=l]{90}{\scriptsize{Pets~\cite{parkhi12a}}} & 
        \rotatebox[origin=l]{90}{\scriptsize{Rendered SST2~\cite{clip}}} & 
        \rotatebox[origin=l]{90}{\scriptsize{RESISC45~\cite{cheng2017remote}}} & 
        \rotatebox[origin=l]{90}{\scriptsize{STL-10~\cite{coates2011analysis}}} & 
        \rotatebox[origin=l]{90}{\scriptsize{VOC2007~\cite{pascal-voc-2007}}} &
        \rotatebox[origin=l]{90}{\ph{.}\textbf{avg. top-1 acc.}}
        \\
        \shline

        \rgray
        \scriptsize \evaOneclip-g/14\phgray{+} & \scriptsize 78.5 & \scriptsize 71.5 & \scriptsize 73.6 & \scriptsize 92.5 & \scriptsize 67.6 & \scriptsize 72.3 & \scriptsize 98.3 & \scriptsize 88.7 & \scriptsize 62.6 & \scriptsize 87.7 & \scriptsize 74.2 & \scriptsize 32.4 & \scriptsize 28.9 & \scriptsize 91.7 & \scriptsize 65.8 & \scriptsize 61.7 & \scriptsize 73.8 & \scriptsize 52.2 & \scriptsize 74.5 & \scriptsize 93.5 & \scriptsize 49.3 & \scriptsize 49.9 & \scriptsize 94.2 & \scriptsize 58.4 & \scriptsize 70.3 & \scriptsize 98.9 & \scriptsize 85.7 & \scriptsize 72.2 \\
        
        \rgray
        \scriptsize \evaOneclip-g/14+ & \scriptsize 79.3 & \scriptsize 72.5 & \scriptsize 74.1 & \scriptsize 92.7 & \scriptsize 68.4 & \scriptsize 75.3 & \scriptsize 99.1 & \scriptsize 90.1 & \scriptsize 72.0 & \scriptsize 89.5 & \scriptsize 74.7 & \scriptsize 39.9 & \scriptsize 31.8 & \scriptsize 90.7 & \scriptsize 70.2 & \scriptsize 67.8 & \scriptsize 73.2 & \scriptsize 56.0 & \scriptsize 79.7 & \scriptsize 93.7 & \scriptsize 66.5 & \scriptsize 62.4 & \scriptsize 94.9 & \scriptsize 58.6 & \scriptsize 71.4 & \scriptsize \underline{99.5} & \scriptsize 84.7 & \scriptsize 75.1 \\
        
        \scriptsize OpenCLIP-G/14\ph{+} & \scriptsize 80.4 & \scriptsize 73.6 & \scriptsize 69.3 & \scriptsize 92.8 & \scriptsize 69.9 & \scriptsize 73.0 & \scriptsize 98.3 & \scriptsize 87.5 & \scriptsize 71.6 & \scriptsize 89.4 & \scriptsize 75.0 & \scriptsize 53.6 & \scriptsize 34.9 & \scriptsize 94.9 & \scriptsize 73.0 & \scriptsize 69.1 & \scriptsize 71.1 & \scriptsize \textbf{59.6} & \scriptsize 81.5 & \scriptsize 93.1 & \scriptsize 62.7 & \scriptsize 63.6 & \scriptsize 95.3 & \scriptsize 65.3 & \scriptsize 72.6 & \scriptsize 98.5 & \scriptsize \underline{87.4} & \scriptsize 76.2 \\

        \scriptsize InternVL-C \ph{+} & \scriptsize 83.2 & \scriptsize 77.3 & \scriptsize {83.8} & \scriptsize \textbf{95.7} & \scriptsize \underline{74.3} & \scriptsize 80.6 & \scriptsize \textbf{99.4} & \scriptsize 93.1 & \scriptsize 80.6 & \scriptsize 89.5 & \scriptsize 76.3 & \scriptsize 53.3 & \scriptsize 35.1 & \scriptsize 94.4 & \scriptsize 69.2 & \scriptsize 70.8 & \scriptsize \underline{79.4} & \scriptsize 56.2 & \scriptsize 85.8 & \scriptsize 95.3 & \scriptsize 65.5 & \scriptsize 48.7 & \scriptsize 96.3 & \scriptsize \textbf{68.4} & \scriptsize 74.4 & \scriptsize 99.4 & \scriptsize 80.0 & \scriptsize 78.0 \\

        \scriptsize DFN5B-CLIP-H/14 \ph{+} & \scriptsize 83.5 & \scriptsize 77.4 & \scriptsize {71.7} & \scriptsize 92.9 & \scriptsize 72.8 & \scriptsize 76.7 & \scriptsize 98.8 & \scriptsize 90.5 & \scriptsize \textbf{85.8} & \scriptsize 89.5 & \scriptsize 77.0 & \scriptsize \underline{71.4} & \scriptsize 34.4 & \scriptsize \underline{95.8} & \scriptsize 77.4 & \scriptsize 70.7 & \scriptsize 65.2 & \scriptsize 54.7 & \scriptsize \textbf{92.5} & \scriptsize \underline{95.8} & \scriptsize 67.7 & \scriptsize \underline{65.2} & \scriptsize \underline{96.5} & \scriptsize 54.8 & \scriptsize \underline{76.1} & \scriptsize 98.9 & \scriptsize 81.5 & \scriptsize 78.3 \\
        
        \rgray
        \scriptsize \evaTwoclip-E/14+ & \scriptsize 82.1 & \scriptsize 75.7 & \scriptsize {82.1} & \scriptsize 94.7 & \scriptsize 72.2 & \scriptsize 79.6 & \scriptsize \underline{99.3} & \scriptsize \underline{93.2} & \scriptsize 74.7 & \scriptsize \textbf{90.5} & \scriptsize 75.3 & \scriptsize 58.7 & \scriptsize 37.0 & \scriptsize 94.7 & \scriptsize 77.6 & \scriptsize 68.2 & \scriptsize 75.9 & \scriptsize 59.0 & \scriptsize 84.5 & \scriptsize 94.9 & \scriptsize 67.7 & \scriptsize 64.4 & \scriptsize 96.0 & \scriptsize 62.6 & \scriptsize 75.7 & \scriptsize 99.3 & \scriptsize \textbf{87.9} & \scriptsize 78.7 \\

        \scriptsize DFN5B-CLIP-H/14+ & \scriptsize \textbf{84.3} & \scriptsize \textbf{78.3} & \scriptsize {79.6} & \scriptsize 93.6 & \scriptsize 73.3 & \scriptsize 79.6 & \scriptsize 98.8 & \scriptsize 90.5 & \scriptsize 83.6 & \scriptsize 88.9 & \scriptsize \underline{77.4} & \scriptsize \textbf{72.5} & \scriptsize 37.9 & \scriptsize \textbf{96.0} & \scriptsize \textbf{80.5} & \scriptsize 70.9 & \scriptsize 61.1 & \scriptsize 56.1 & \scriptsize \underline{91.6} & \scriptsize \textbf{96.2} & \scriptsize 67.9 & \scriptsize \textbf{69.6} & \scriptsize \textbf{96.8} & \scriptsize 55.5 & \scriptsize 75.9 & \scriptsize 99.1 & \scriptsize 81.9 & \scriptsize 79.2 \\
        
        \rgray
        \scriptsize \evaclipeight\ph{+} & \scriptsize 83.5 & \scriptsize 77.7 & \scriptsize \underline{85.2} & \scriptsize \underline{95.3} & \scriptsize \underline{74.3} & \scriptsize \underline{81.2} & \scriptsize \underline{99.3} & \scriptsize 92.3 & \scriptsize \underline{84.8} & \scriptsize 89.6 & \scriptsize 76.2 & \scriptsize 60.5 & \scriptsize \underline{41.7} & \scriptsize 94.8 & \scriptsize 79.0 & \scriptsize \underline{71.0} & \scriptsize 68.9 & \scriptsize 56.1 & \scriptsize 86.4 & \scriptsize 95.5 & \scriptsize \underline{70.9} & \scriptsize 58.1 & \scriptsize 96.4 & \scriptsize 66.2 & \scriptsize 75.3 & \scriptsize 99.3 & \scriptsize 85.1 & \scriptsize \underline{79.4} \\

        \rgray
        \scriptsize \evaclipx \phgray{+} & \scriptsize \underline{83.8} & \scriptsize \underline{77.9} & \scriptsize \textbf{87.3} & \scriptsize \textbf{95.7} & \scriptsize \textbf{74.7} & \scriptsize \textbf{82.2} & \scriptsize \textbf{99.4} & \scriptsize \textbf{93.8} & \scriptsize 83.0 & \scriptsize \underline{89.8} & \scriptsize \textbf{77.7} & \scriptsize 59.7 & \scriptsize \textbf{43.1} & \scriptsize 94.9 & \scriptsize \underline{79.9} & \scriptsize \textbf{72.1} & \scriptsize \textbf{79.8} & \scriptsize \underline{59.3} & \scriptsize 86.0 & \scriptsize \underline{95.8} & \scriptsize \textbf{72.4} & \scriptsize \underline{65.2} & \scriptsize 96.1 & \scriptsize \underline{67.5} & \scriptsize \textbf{76.9} & \scriptsize \textbf{99.6} & \scriptsize 85.8 & \scriptsize \textbf{80.7} \\

        \end{tabular}
\caption{\textbf{\evaclip zero-shot image classification performance on 27 datasets.} We report top-1 accuracy on all datasets. The best results are in \textbf{bold} and the second best are \underline{underlined}.}
\label{tab: clip zs img cls 27}
\end{table*}

This paper introduces \evaclipx, the largest open-source CLIP model with 18-billion parameters to narrow this gap.
\evaclip~\cite{sun2023evaclip} open-sources a series of effective and efficient CLIP models, which have been leveraged as the vision foundation by numerous impactful works across 2D / 3D vision and multimodal modeling~\cite{blip2,zhu2023minigpt4,uni3d,pan2023tap,wang2023cogvlm,emu1}. We further scale up \evaclip to this significant parameter size building upon the scaling philosophy of \evaone~\cite{eva,EVA02} and \evaclip~\cite{sun2023evaclip}.
With merely 6-billion training samples seen and trained on publicly available datasets, \evaclipx achieves the exceptional \textbf{80.7}\% average zero-shot top-1 accuracy on 27 widely recognized image classification benchmarks, significantly surpassing its forerunner \evaTwoclip-E/14+ (5-billion parameters) and other open-source CLIP models. 
Besides, the models \textit{have not exhibited any signal of performance saturation}, shedding light on further scaling of vision models.
An intuitive demonstration is shown in \Cref{fig:teaser}.

The successful training of \evaclipx exemplifies the potential of the EVA-style visual model scaling philosophy. We keep open-sourcing the training code and weights of our models to encourage further research and empower the development of vision and multimodal foundation models.


\begin{table*}[h]
\centering
    \tablestyle{4.2pt}{1.2}
    \begin{tabular}{l|ccc|ccc|ccc|ccc|c}
        & \multicolumn{6}{c|}{zero-shot \textbf{text} retrieval} & \multicolumn{6}{c|}{zero-shot \textbf{image} retrieval} \\
        & \multicolumn{3}{c|}{\scriptsize Flickr30K} & \multicolumn{3}{c|}{\scriptsize COCO} & \multicolumn{3}{c|}{\scriptsize Flickr30K} & \multicolumn{3}{c|}{\scriptsize COCO} \\
    
        method{\scriptsize{\ph{+}}} & \scriptsize R@1 & \scriptsize R@5 & \scriptsize R@10 & \scriptsize R@1 & \scriptsize R@5 & \scriptsize R@10 & \scriptsize R@1 & \scriptsize R@5 & \scriptsize R@10 & \scriptsize R@1 & \scriptsize R@5 & \scriptsize R@10 & MR \\
        \shline

        \rgray
        \scriptsize \evaOneclip-g/14 \phgray{+} & \scriptsize 87.9 & \scriptsize 98.0 & \scriptsize 99.5 & \scriptsize 61.7 & \scriptsize 83.2 & \scriptsize 89.9 & \scriptsize 72.5 & \scriptsize 91.5 & \scriptsize 95.4 & \scriptsize 44.5 & \scriptsize 69.1 & \scriptsize 77.7 & \scriptsize 80.9 \\
        
        \rgray
        \scriptsize \evaOneclip-g/14+ & \scriptsize 93.3 & \scriptsize 99.5 & \scriptsize 99.9 & \scriptsize 69.4 & \scriptsize 88.3 & \scriptsize 93.2 & \scriptsize 79.2 & \scriptsize 95.2 & \scriptsize 97.3 & \scriptsize 51.1 & \scriptsize 74.7 & \scriptsize 82.5 & \scriptsize 85.3 \\
        
        \scriptsize OpenCLIP-G/14\ph{+} & \scriptsize 93.5 & \scriptsize 99.3 & \scriptsize 99.7 & \scriptsize 69.0 & \scriptsize 87.8 & \scriptsize 93.1 & \scriptsize 80.9 & \scriptsize 95.1 & \scriptsize 97.2 & \scriptsize 52.6 & \scriptsize 76.1 & \scriptsize 83.6 & \scriptsize 85.7 \\
        
        \rgray
        \scriptsize \evaTwoclip-E/14+ & \scriptsize 94.3 & \scriptsize 99.6 & \scriptsize 99.8 & \scriptsize 69.4 & \scriptsize 88.6 & \scriptsize 93.3 & \scriptsize 79.7 & \scriptsize 94.9 & \scriptsize 97.3 & \scriptsize 52.5 & \scriptsize 75.9 & \scriptsize 83.4 & \scriptsize 85.7 \\
        
        \rgray
        \scriptsize \evaclipeight\ph{+} & \scriptsize 95.6 & \scriptsize 99.6 & \scriptsize 99.9 & \scriptsize 70.3 & \scriptsize 89.3 & \scriptsize 93.9 & \scriptsize 80.8 & \scriptsize 95.5 & \scriptsize 97.6 & \scriptsize 53.0 & \scriptsize 76.0 & \scriptsize 83.4 & \scriptsize 86.2 \\

        \scriptsize DFN5B-CLIP-H/14 \ph{+} & \scriptsize 92.9 & \scriptsize 99.3 & \scriptsize 99.9 & \scriptsize 72.3 & \scriptsize 90.2 & \scriptsize 94.6 & \scriptsize 80.1 & \scriptsize 95.2 & \scriptsize 97.3 & \scriptsize 53.9 & \scriptsize 78.0 & \scriptsize 85.6 & \scriptsize 86.6 \\

        \scriptsize InternVL-C \ph{+} & \scriptsize 93.8 & \scriptsize \textbf{99.7} & \scriptsize \textbf{100.0} & \scriptsize 70.3 & \scriptsize 89.2 & \scriptsize 93.8 & \scriptsize 82.1 & \scriptsize 96.0 & \scriptsize 98.1 & \scriptsize 54.1 & \scriptsize 77.1 & \scriptsize 84.8 & \scriptsize 86.6 \\

        \scriptsize DFN5B-CLIP-H/14 + & \scriptsize 93.6 & \scriptsize 99.3 & \scriptsize 99.6 & \scriptsize 71.8 & \scriptsize 90.4 & \scriptsize 94.9 & \scriptsize 82.1 & \scriptsize 96.0 & \scriptsize 97.9 & \scriptsize 55.6 & \scriptsize \textbf{79.2} & \scriptsize \textbf{86.3} & \scriptsize 87.2 \\

        \rgray
        \scriptsize \evaclipx \phgray{+} & \scriptsize \textbf{96.7} & \scriptsize \textbf{99.7} & \scriptsize \textbf{100.0} & \scriptsize \textbf{73.6} & \scriptsize \textbf{90.9} & \scriptsize \textbf{95.0} & \scriptsize \textbf{83.3} & \scriptsize \textbf{96.3} & \scriptsize \textbf{98.3} & \scriptsize \textbf{56.2} & \scriptsize 78.5 & \scriptsize 85.6 & \scriptsize \textbf{87.8} \\

    \end{tabular}
\caption{Zero-shot retrieval performance on Flickr30K~\cite{flickr30K} and COCO~\cite{lin2014coco}.}
\label{tab: clip zs retrieval}
\end{table*}

\section{Weak-to-Strong Vision Scaling}
Our scaling-up procedure is guided by the principles of \evaone~\cite{eva} and \evaclip~\cite{sun2023evaclip}. 
The EVA philosophy for scaling visual models follows a weak-to-strong paradigm, designed to improve visual models through a strategic progression.
This process begins with a large \evaone vision model distilling knowledge from a small \evaclip model, which in turn serves as the vision encoder initialization to stabilize and accelerate the training of a larger \evaclip. After that, the closed-loop scaling-up cycle continues and a larger \evaone is distilled out.
Throughout our model scaling cycle, the training dataset remains largely fixed to demonstrate the effectiveness of our model-scale specific scaling philosophy, although scaling up datasets can further unleash the scaling power of our method.

Specifically, in this work, we pre-train a large \evaone model named \evaone-18B using a small \evaclip (\evaTwoclip-E/14+)~\cite{sun2023evaclip} as the teacher, which is trained to reconstruct masked image-text aligned vision features from \evaTwoclip-E/14+. \evaone-18B omits bias terms of QKV projections and uses RMSNorm~\cite{rmsnorm} instead of LayerNorm~\cite{layernorm} following LLaMA~\cite{touvron2023llama}. Subsequently, we leverage the \evaone model as the vision encoder initialization for \evaclip pre-training with the image-text contrastive learning objective. Besides, we also introduce a smaller counterpart, \evaclipeight, which undergoes similar pre-training methodologies. Notably, our experiments demonstrate sustained performance improvement with the progressive weak-teach-strong scaling up of \evaclip.

\section{Experiments}
\paragraph{Settings.} Following \evaclip~\cite{sun2023evaclip}, we initialized the model with pre-trained vision and text encoders. Specifically, we employ a pre-trained \evaone-18B ~\cite{eva, EVA02} as the vision encoder and \evaTwoclip-E/14+ ~\cite{sun2023evaclip} for the text encoder. We adopt the LAMB optimizer~\cite{lamb} with $\beta_1$ = 0.9, $\beta_2$=0.95, and a weight decay of 0. We apply different learning rates and layer decay rates to the vision encoder and text encoder to ensure optimal training. We set the peak learning rate as 4e-4 and 4e-5 for the vision encoder and the text encoder respectively, with 2000 warm-up steps. Afterwards, the learning rates decay to 0 with a cosine schedule. The learning rate layer decay rates are configured as 0.9 and 0.75 for the vision and text encoders. The temperature parameter remains constant at 0.01. Further, we use the $\mathtt{DeepSpeed}$ optimization library~\cite{rasley2020deepspeed} with ZeRO stage-3 partition~\cite{rajbhandari2020zero}, gradient checkpointing~\cite{gradcheckpointing} and flash attention~\cite{flashattention} to optimize the training cost.

\paragraph{Dataset.} Our Merged-2B dataset consists of 1.6 billion samples from LAION-2B~\cite{laion5b} and 0.4 billion samples from COYO-700M~\cite{kakaobrain2022coyo-700m}. Note that the use of a subset from LAION-2B is not the result of deliberate filtering, but rather due to image downloading failures. The use of 0.4 billion COYO-700M samples aims to complement the number of training samples to nearly the same as LAION-2B.
Merged-2B+ consists of all samples from Merged-2B, along with additional 20 million samples from LAION-COCO~\cite{laioncoco} and 23 million samples from Merged-video including VideoCC~\cite{videocc}, InternVid~\cite{wang2024internvid} and WebVid-10M~\cite{bain2021frozen}. Merged-video is included at the end of the training process.

\evaclipx pre-trains with 5.4 billion samples from Merged-2B seen with 50\% of patch dropout ratio~\cite{flip}, 0.6 billion samples from Merged-2B and 20 million samples from LAION-COCO without patch dropout, and 24 million samples from Merged-video with 50\% of patch dropout ratio.


\begin{table}[t!]
    \tablestyle{0.8pt}{1.2}
    \scriptsize
    \begin{tabular}{l|ccc|ccc|ccc}
        & \multicolumn{3}{c|}{image encoder} & \multicolumn{3}{c|}{text encoder} & \multicolumn{3}{c}{\# \scriptsize{params}} \\
        method & layers & width & heads & layers & width & heads & image & text & total \\ 
        \shline
        
        \evaclipeight\ph{+} & 32 & 4096 & 32 & 32 & 1280 & 20 & 7.5B & 695M & 8.1B \\

        \evaclipx \ph{+} & 48 & 5120 & 40 & 32 & 1280 & 20 & 17.5B & 695M & 18.1B \\

    \end{tabular}    
    \caption{Architecture configurations.
    }
    \label{evacliparch}
\end{table}

\paragraph{Evaluation.} We evaluate on 33 widely used datasets across image, video classification and image-text retrieval.
All datasets used to evaluate \evaclipx are reported in \tblref{tab: eval_dataset_table}. 
We utilize the specified prompt templates following~\cite{clip, openclip}.

\paragraph{Zero-Shot Image Classification.} We show the exceptional performance of \evaclip on all 27 zero-shot image classification benchmarks in~\tblref{tab: clip zs img cls 27}. \evaclipx achieves \textbf{80.7\%} top-1 accuracy averaged across all 27 benchmarks. These results significantly outperform the previous best open-sourced DFN5B-CLIP-H/14+~\cite{dfn5b} by \evagreen{\textbf{+1.5\%}}, and the largest existing CLIP model, InternVL-C~\cite{chen2023internvl}, by \evagreen{\textbf{+2.7\%}}. For Birdsnap dataset, the download was limited to 2195 test images due to broken links.

\begin{table}[h]
\centering
    \tablestyle{1.5pt}{1.2}
    \begin{tabular}{l|c|cccc|c}
        method\ph{+} & \#Frames & \scriptsize UCF-101 & \scriptsize K-400  & \scriptsize K-600 & \scriptsize K-700 & \textbf{avg.} \\
        \shline

        \rgray
        \scriptsize \evaOneclip-g/14\phgray{+} & 1 & \scriptsize 76.0 & \scriptsize 65.4 & \scriptsize 64.5 & \scriptsize 58.8 & \scriptsize 66.2 \\
        
        \scriptsize DFN5B-CLIP-H/14\ph{+} & 1 & \scriptsize 78.2 & \scriptsize 65.2 & \scriptsize 65.5 & \scriptsize 59.2 & \scriptsize 67.0 \\

        \scriptsize DFN5B-CLIP-H/14+ & 1 & \scriptsize 79.2 & \scriptsize 66.7 & \scriptsize 67.0 & \scriptsize 60.7 & \scriptsize 68.4 \\
        
        \scriptsize OpenCLIP-G/14\ph{+} & 1 & \scriptsize 80.5 & \scriptsize 67.1 & \scriptsize 66.9 & \scriptsize 60.3 & \scriptsize 68.7 \\
        
        \rgray
        \scriptsize \evaOneclip-g/14+ & 1 & \scriptsize 78.9 & \scriptsize 67.3 & \scriptsize 67.3 & \scriptsize 61.5 & \scriptsize 68.8 \\
        
        \rgray
        \scriptsize \evaTwoclip-E/14+ & 1 & \scriptsize 83.1 & \scriptsize 70.7 & \scriptsize 70.0 & \scriptsize 64.4 & \scriptsize 72.1 \\
        
        \rgray
        \scriptsize \evaclipeight\phgray{+} & 1 & \scriptsize 85.7 & \scriptsize 71.3 & \scriptsize 71.2 & \scriptsize 66.1 & \scriptsize 73.6 \\

        \scriptsize InternVL-C \ph{+} & 1 & \scriptsize 85.2 & \scriptsize 71.8 & \scriptsize 71.7 & \scriptsize 66.4 & \scriptsize 73.7 \\

        \rgray
        \scriptsize \evaclipx \phgray{+} & 1 & \scriptsize \textbf{86.0} & \scriptsize \textbf{72.9} & \scriptsize \textbf{72.9} & \scriptsize \textbf{68.2} & \scriptsize \textbf{75.0} \\

        \hline

        \rgray
        \scriptsize \evaclipx \phgray{+} & 8 & \scriptsize \textbf{88.2} & \scriptsize \textbf{79.3} & \scriptsize \textbf{79.2} & \scriptsize \textbf{72.1} & \scriptsize \textbf{79.7} \\

        \rgray
        \scriptsize \evaclipx \phgray{+} & 16 & \scriptsize \textbf{88.4} & \scriptsize \textbf{79.4} & \scriptsize \textbf{79.4} & \scriptsize \textbf{72.2} & \scriptsize \textbf{79.8} \\

    \end{tabular}
\caption{\textbf{\evaclip zero-shot video classification performance.} We report top1 accuracy for UCF-101~\cite{ucf101}, average of top1 and top5 accuracy for Kinetics-400~\cite{carreira2017quo}, Kinetics-600~\cite{k600} and Kinetics-700~\cite{k700}.}
\label{tab: clip zs video cls 4}
\end{table}

\begin{table*}[t!]
\vspace{-10pt}
\centering
\subfloat[
\textbf{Zero-shot performance on ImageNet variants and ObjectNet.} ``avg. acc.'': the averaged top-1 accuracy on different ImageNet variants (\ie, IN-\{1K, V2, ReaL, Adv., Ren., Ske.\}), and ObjectNet. ``{$\Delta$\scriptsize{$\downarrow$}}'': The gap between the averaged top-1 accuracy and the ImageNet-1K top-1 accuracy (the lower the better). \evaclip suffers from the smallest performance drop (only \textbf{\evagreen{0.2}}\% top-1 accuracy gap for \evaclipx) while \evaclipx achieves \textbf{\evagreen{83.6}}\% top-1 accuracy averaged on all 6 benchmarks.
\label{tab: clip result}
]{
\centering
\begin{minipage}{1\linewidth}{\begin{center}
\tablestyle{3.5pt}{1.2}
    \begin{tabular}{l|cccccc|c|c}
        method\ph{+} & \scriptsize IN-1K & \scriptsize IN-A & \scriptsize IN-R & \scriptsize IN-V2 & \scriptsize IN-Sketch & \scriptsize ObjectNet & \textbf{$\Delta$\scriptsize{$\downarrow$}} & avg. acc. \\

        \shline

        \scriptsize DFN5B-CLIP-H/14\ph{+} & \scriptsize 83.5 & \scriptsize 71.7 & \scriptsize 92.9 & \scriptsize 77.4 & \scriptsize 72.8 & \scriptsize 76.7 & \scriptsize 4.4 & \scriptsize 79.2\\

        \scriptsize OpenCLIP-G/14\ph{+} & \scriptsize 80.4 & \scriptsize 69.3 & \scriptsize 92.8 & \scriptsize 73.6 & \scriptsize 69.9 & \scriptsize 73.0 & \scriptsize{3.9} & \scriptsize 76.5\\
        
        \scriptsize SigLIP-SO~\cite{siglip}~(reported) & \scriptsize 82.0 & \scriptsize 71.9 & \scriptsize 95.1 & \scriptsize 76.1 & \scriptsize 74.0 & \scriptsize 70.6 & \scriptsize 3.7 & \scriptsize 78.3 \\

        \scriptsize DFN5B-CLIP-H/14+ & \scriptsize 84.3 & \scriptsize 79.6 & \scriptsize 93.6 & \scriptsize 78.3 & \scriptsize 73.3 & \scriptsize 79.6 & \scriptsize 2.8 & \scriptsize 81.5 \\

        \rgray
        \scriptsize \evaOneclip-g/14 \phgray{+} & \scriptsize 78.5 & \scriptsize 73.6 & \scriptsize 92.5 & \scriptsize 71.5 & \scriptsize 67.6 & \scriptsize 72.3 & \scriptsize{2.5} & \scriptsize 76.0 \\
        
        \rgray
        \scriptsize \evaOneclip-g/14+ & \scriptsize 79.3 & \scriptsize 74.1 & \scriptsize 92.7 & \scriptsize 72.5 & \scriptsize 68.4 & \scriptsize 75.3 & \scriptsize{2.2} & \scriptsize 77.1\\

        \scriptsize BASIC-L~\cite{basic}~(reported) & \scriptsize 85.7 & \scriptsize 85.6 & \scriptsize 95.7 & \scriptsize 80.6 & \scriptsize 76.1 & \scriptsize 82.3 & \scriptsize 1.4 & \scriptsize 84.3 \\

        \scriptsize SigLIP-SO+~\cite{siglip}~(reported) & \scriptsize 83.0 & \scriptsize 82.5 & \scriptsize 95.8 & \scriptsize 77.2 & \scriptsize 74.5 & \scriptsize 77.0 & \scriptsize 1.3 & \scriptsize 81.7 \\

        \rgray
        \scriptsize \evaTwoclip-E/14+ & \scriptsize 82.1 & \scriptsize 82.1 & \scriptsize 94.7 & \scriptsize 75.7 & \scriptsize 72.2 & \scriptsize 79.6 & \scriptsize 1.0 & \scriptsize 81.1 \\

        \scriptsize InternVL-C \ph{+} & \scriptsize 83.2 & \scriptsize 83.8 & \scriptsize 95.7 & \scriptsize 77.3 & \scriptsize 74.3 & \scriptsize 80.6 & \scriptsize 0.7 & \scriptsize 82.5 \\
        
        \rgray
        \scriptsize \evaclipeight \phgray{+} & \scriptsize 83.5 & \scriptsize 85.2 & \scriptsize 95.3 & \scriptsize 77.7 & \scriptsize 74.3 & \scriptsize 81.2 & \scriptsize 0.6 & \scriptsize 82.9 \\

        \rgray
        \scriptsize \evaclipx \phgray{+} & \scriptsize 83.8 & \scriptsize 87.3 & \scriptsize 95.7 & \scriptsize 77.9 & \scriptsize 74.7 & \scriptsize 82.2 & \scriptsize \textbf{0.2} & \scriptsize 83.6

    \end{tabular}
\end{center}}\end{minipage}
}
\\
\subfloat[
{\textbf{Comparison \evaclipx's zero-shot image classification performance with BASIC-L~\cite{basic} on 17 datasets.} Our report includes the top-1 accuracy for all datasets, considering that BASIC-L only provided top-1 accuracy for these specific 17 datasets. ( ) is the average top-1 accuracy removing Birdsnap due to the different test size between \evaclipx and BASIC-L. \evaclipx outperforms BASIC-L with a notable margin of \textbf{\evagreen{+8.2}}~(\textbf{\evagreen{+6.3}}) in average top-1 accuracy, despite exhibiting lower performance on ImageNet variants.}
\label{tab: clip robustness}
]{
\centering
\begin{minipage}{1\linewidth}{\begin{center}
    \tablestyle{3.5pt}{1.2}
    \begin{tabular}{l|cccccc|ccccccccccc|c}
        \scriptsize method &
        \rotatebox[origin=l]{90}{\scriptsize{ImageNet-1K~\cite{deng2009imagenet}}} &
        \rotatebox[origin=l]{90}{\scriptsize{ImageNet-V2~\cite{recht2019imagenetv2}}} &
        \rotatebox[origin=l]{90}{\scriptsize{ImageNet-Adv.~\cite{inadv}}} &
        \rotatebox[origin=l]{90}{\scriptsize{ImageNet-Ren.~\cite{inren}}} &
        \rotatebox[origin=l]{90}{\scriptsize{ImageNet-Ske.~\cite{inske}}} &
        \rotatebox[origin=l]{90}{\scriptsize{ObjectNet~\cite{objectnet}}} &
        \rotatebox[origin=l]{90}{\scriptsize{CIFAR-10~\cite{cifar}}} &
        \rotatebox[origin=l]{90}{\scriptsize{CIFAR-100~\cite{cifar}}} & 
        \rotatebox[origin=l]{90}{\scriptsize{MNIST~\cite{lecun1998gradient}}} & 
        \rotatebox[origin=l]{90}{\scriptsize{SUN397~\cite{xiao2010sun}}} &
        \rotatebox[origin=l]{90}{\scriptsize{Birdsnap~\cite{berg2014birdsnap}}} & 
        \rotatebox[origin=l]{90}{\scriptsize{DTD~\cite{cimpoi14describing}}} & 
        \rotatebox[origin=l]{90}{\scriptsize{EuroSAT~\cite{helber2019eurosat}}} &

        \rotatebox[origin=l]{90}{\scriptsize{Food-101~\cite{bossard2014food}}} & 
        
        \rotatebox[origin=l]{90}{\scriptsize{PCam~\cite{veeling2018rotation}}} & 
        
        \rotatebox[origin=l]{90}{\scriptsize{RESISC45~\cite{cheng2017remote}}} & 
        \rotatebox[origin=l]{90}{\scriptsize{STL-10~\cite{coates2011analysis}}} & 
        
        \rotatebox[origin=l]{90}{\ph{.}\textbf{avg. top-1 acc.}}
        \\
        \shline

        \scriptsize BASIC-L~\cite{basic}~(reported) & \scriptsize 85.7 & \scriptsize 80.6 & \scriptsize 85.6 & \scriptsize 95.7 & \scriptsize 76.1 & \scriptsize 82.3 & \scriptsize 97.5 & \scriptsize 82.3 & \scriptsize 40.3 & \scriptsize 76.2 & \scriptsize 59.2 & \scriptsize 64.6 & \scriptsize 51.0 & \scriptsize 95.1 & \scriptsize 59.6 & \scriptsize 72.7 & \scriptsize 99.6 & \scriptsize 76.7~(77.8) \\

        \rgray
        \scriptsize \evaclipx \phgray{+} & \scriptsize 83.8 & \scriptsize 77.9 & \scriptsize 87.3 & \scriptsize 95.7 & \scriptsize 74.7 & \scriptsize 82.2 & \scriptsize 99.4 & \scriptsize 93.8 & \scriptsize 83.0 & \scriptsize 77.7 & \scriptsize 79.9 & \scriptsize 72.1 & \scriptsize 79.8 & \scriptsize 95.8 & \scriptsize 65.2 & \scriptsize 76.9 & \scriptsize 99.6 & \scriptsize 84.9~(84.1) \\

        \rgray
        \scriptsize  & \scriptsize \textbf{\evared{-1.9}} & \scriptsize \textbf{\evared{-2.7}} & \scriptsize \textbf{\evagreen{+1.7}} & \scriptsize +0.0 & \scriptsize \textbf{\evared{-1.4}} & \scriptsize \textbf{\evared{-0.1}} & \scriptsize \textbf{\evagreen{+1.9}} & \scriptsize  \textbf{\evagreen{+11.5}} & \scriptsize \textbf{\evagreen{+42.7}} & \scriptsize \textbf{\evagreen{+1.5}} & \scriptsize \textbf{\evagreen{+20.7}} & \scriptsize \textbf{\evagreen{+7.5}} & \scriptsize \textbf{\evagreen{+28.8}} & \scriptsize \textbf{\evagreen{+0.7}} & \scriptsize \textbf{\evagreen{+5.6}} & \scriptsize \textbf{\evagreen{+4.2}} & \scriptsize +0.0 & \scriptsize \textbf{\evagreen{+8.2}}~(\textbf{\evagreen{+6.3}}) \\
                
        \end{tabular}
\end{center}}\end{minipage}
}

\caption{\textbf{Robustness evaluation of CLIP models and comparison with BASIC-L~\cite{basic} on 17 Benchmarks.}}
\label{tab: robustness and imagenet}
\end{table*}

\paragraph{Zero-Shot Video Classification.} We report the top-1 accuracy for UCF-101~\cite{ucf101} and the mean of top-1 and top-5 accuracy for Kinetics-400~\cite{carreira2017quo}, Kinetics-600~\cite{k600} and Kinetics-700~\cite{k700}. In~\tblref{tab: clip zs video cls 4} we demonstrate that \evaclipx also outperforms other CLIP models on zero-shot video classification benchmarks by a large margin. When sampling a single center frame per video, \evaclipx achieves accuracies of 86.0\%, 72.9\%, 72.9\%, and 68.2\% across the four evaluated benchmarks. Further, when uniformly sample 8 or 16 frames per video, we observe an improvement of \evagreen{\textbf{+4.7\%}} / \evagreen{\textbf{+4.8\%}} averaged across four benchmarks compared to the single-frame setting.

\paragraph{Zero-Shot Image-Text Retrieval.} In \tblref{tab: clip zs retrieval}, we report the zero-shot image and text retrieval results on Flickr30K~\cite{flickr30K} and COCO~\cite{lin2014coco}. \evaclipx achieves an average recall of 87.8\% across all retrieval benchmarks, significantly outperforming competitors.

\paragraph{Robustness.} In~\tblref{tab: robustness and imagenet}, we demonstrate that scaling up \evaclip significantly enhances the robustness of visual representations. \evaclip suffers from the smallest performance drop ({$\Delta$\scriptsize{$\downarrow$}}) between ImageNet-1K and ImageNet variants including adversarial ones, with merely \textbf{\evagreen{0.2}}\% top-1 accuracy gap for \evaclipx. 

For a more robust and comprehensive evaluation of robustness and zero-shot performance, it is advisable to include more benchmarks covering more image distributions.
However, we want to note that higher ImageNet top-1 accuracy does not necessarily lead to better overall performance, as evidenced in \tblref{tab: clip robustness}, where BASIC-L~\cite{basic} exhibits higher ImageNet-related top-1 accuracy but considerably lower overall average top-1 accuracy compared to \evaclipx across a broader range of datasets and distributions, showing a difference of -8.2\%.

\paragraph{Linear Probing on ImageNet-1K.} In \tblref{tab: linear_prob_in1k}, we present the results of linear probing on ImageNet-1K~\cite{deng2009imagenet}. \evaclipx achieves an average top-1 accuracy of 88.9\%, surpassing InternVL-C~\cite{chen2023internvl} by 0.7\%.


\begin{table}[h]
\centering
    \tablestyle{1.5pt}{1.2}
    \begin{tabular}{l|c|c}
        \scriptsize method & \scriptsize \#param & \scriptsize top1 acc. \\
        \shline
        
         \scriptsize OpenCLIP-G/14~(reported) & \scriptsize 1.8B & \scriptsize 86.2  \\
         \rgray
         \scriptsize \evaOneclip-g/14 \ph{+} & \scriptsize 1.0B & \scriptsize 86.5  \\
         \rgray
         \scriptsize \evaTwoclip-E/14+ & \scriptsize 4.4B & \scriptsize 88.1  \\
         \scriptsize InternVL-C~(reported) & \scriptsize 5.9B & \scriptsize 88.2  \\
         \rgray
         \scriptsize \evaclipeight \ph{+} & \scriptsize 7.5B & \scriptsize 88.5  \\
         \rgray
         \scriptsize \evaclipx \ph{+} & \scriptsize 17.5B & \scriptsize 88.9 \\

    \end{tabular}
\caption{\textbf{Linear Probing on ImageNet-1K~\cite{deng2009imagenet}.} The top-1 accuracy shows a continuous improvement with the scaling up of \evaclip.}

\label{tab: linear_prob_in1k}
\end{table}

\paragraph{3D Representation.} We adopt the Uni3D~\cite{uni3d} setting to explore the effectiveness of scaling up teachers.
With the scaling up of \evaclip in \tblref{tab: 3d_representation}, we observe consistent improvements in 3D representation learning capabilities. Further, Uni3D-base equipped with \evaclipx sets new records on ModelNet~\cite{wu20153d} and ScanObjectNN~\cite{uy2019revisiting} benchmarks.

\begin{table}[h]
\centering
    \tablestyle{1.5pt}{1.2}
    \begin{tabular}{l|c|ccc}
        \scriptsize teacher  & \scriptsize data & \scriptsize   O-LVIS & \scriptsize MNet40 & \scriptsize  ScanObjNN   \\
        \shline
        
         \scriptsize OpenCLIP-G/14 \ph{+}  & \scriptsize  w/o LVIS & \scriptsize 44.5 & \scriptsize 85.8 & \scriptsize 58.9  \\
         \rgray
         \scriptsize \evaTwoclip-E/14+  & \scriptsize w/o LVIS & \scriptsize  45.8 & \scriptsize 86.1 & \scriptsize 61.7  \\
         \rgray
         \scriptsize \evaclipeight \ph{+}  & \scriptsize w/o LVIS & \scriptsize 46.2 & \scriptsize 87.3 & \scriptsize 62.7    \\
         \rgray
         \scriptsize \evaclipx \ph{+}  & \scriptsize w/o LVIS & \scriptsize \textbf{47.0} & \scriptsize \textbf{87.6} & \scriptsize \textbf{65.3} \\
         \hline
         \rgray
         \scriptsize \evaTwoclip-E/14+  & \scriptsize Ensembled & \scriptsize  51.7 & \scriptsize 86.3 & \scriptsize 63.8  \\
         \rgray
         \scriptsize \evaclipx \ph{+} & \scriptsize Ensembled & \scriptsize \textbf{53.2(\evagreen{+1.5})} & \scriptsize \textbf{88.6(\evagreen{+2.3})} & \scriptsize \textbf{67.8(\evagreen{+4.0})} \\
    \end{tabular}
\caption{\textbf{\evaclipx enhances zero-shot 3d classification performance.} We use Uni3D-base~\cite{uni3d} as the baseline and scale the teacher from 5B to 18B. We report top-1 accuracy on Objaverse-LVIS~\cite{deitke2023objaverse}, ModelNet40~\cite{wu20153d} and ScanObjectNN~\cite{uy2019revisiting}.}

\label{tab: 3d_representation}
\end{table}

\section{Ablation Studies}

\paragraph{Video Data.} In \tblref{tab: ablations_video_data}, we conduct ablations on \evaclipx's zero-shot performance, comparing results when trained with and without Merged-Video. The training objective for the video data aligns with that of images, encompassing the extraction of features from video where 8 frames are uniformly sampled. The mean of all [CLS] embeddings serves as a representation for the video. The outcomes reveal substantial performance improvements associated with training using Merged-Video. The zero-shot performance, averaged across UCF-101~\cite{ucf101} and Kinetics-400~\cite{carreira2017quo} / 600~\cite{k600} / 700~\cite{k700}, indicates a gain of \textbf{\evagreen{+0.7}} for evaluation with one middle frame and \textbf{\evagreen{+0.8}} for evaluation with 8 frames.


\begin{table}[h]
\centering
    \tablestyle{1.5pt}{1.2}
    \begin{tabular}{l|ccc|c}
        & \multicolumn{3}{c|}{classification} & retrieval \\
         & \scriptsize image & \scriptsize video~(\#F 1) & video~(\#F 8) & avg. recall \\
        \shline
        
        w/o video data & \scriptsize 80.7 & \scriptsize 74.3 & \scriptsize 78.9 & \scriptsize 87.9 \\

        w/ video data & \scriptsize 80.7 & \scriptsize 75.0~(\textbf{\evagreen{+0.7}}) & \scriptsize 79.7~(\textbf{\evagreen{+0.8}}) & \scriptsize 87.8~(\textbf{\evared{-0.1}}) \\

    \end{tabular}

\vspace{-1.em}
\caption{\textbf{Video data enhances zero-shot video classification performance}. We respectively report performances averaged on 27 image classification benchmarks, 4 video benchmarks and 2 image-text retrieval benchmarks.}
\label{tab: ablations_video_data}
\end{table}


\begin{table}[t]
\centering
    \tablestyle{1.5pt}{1.2}
    \begin{tabular}{l|c|cccccc|c}
        method\ph{+} & resolution & \scriptsize IN-1K & \scriptsize IN-A & \scriptsize IN-R & \scriptsize IN-V2 & \scriptsize IN-Ske. & \scriptsize ObjectNet & \textbf{avg.} \\

        \shline
        
        \rgray
        \scriptsize \evaclipeight \phgray{+} & 224$\times$224 &\scriptsize 83.5 & \scriptsize 85.2 & \scriptsize 95.3 & \scriptsize 77.7 & \scriptsize 74.3 & \scriptsize 81.2 & \scriptsize 82.9 \\

        \rgray
        \scriptsize \evaclipeight+ & 448$\times$448 & \scriptsize 83.8 & \scriptsize 88.7 & \scriptsize 95.4 & \scriptsize 77.7 & \scriptsize 74.1 & \scriptsize 82.9 & \scriptsize 83.8 \\
        \rgray
         &  & \scriptsize \textbf{\evagreen{+0.3}} & \scriptsize \textbf{\evagreen{+3.5}} & \scriptsize \textbf{\evagreen{+0.1}} & \scriptsize +0.0 & \scriptsize \textbf{\evared{-0.2}} & \scriptsize \textbf{\evagreen{+1.7}} & \scriptsize \textbf{\evagreen{+0.9}} \\

        \midrule

        \rgray
        \scriptsize \evaclipx \phgray{+} & 224$\times$224 & \scriptsize 83.8 & \scriptsize 87.3 & \scriptsize 95.7 & \scriptsize 77.9 & \scriptsize 74.7 & \scriptsize 82.2 & \scriptsize 83.6 \\

        \rgray
        \scriptsize {\evaclipx}+ & 336$\times$336 & \scriptsize 83.9 & \scriptsize 88.9 & \scriptsize 95.6 & \scriptsize 78.2 & \scriptsize 74.3 & \scriptsize 83.6 & \scriptsize 84.1 \\
        \rgray
        & & \scriptsize \textbf{\evagreen{+0.1}} & \scriptsize \textbf{\evagreen{+1.6}} & \scriptsize \textbf{\evared{-0.1}} & \scriptsize \textbf{\evagreen{+0.3}} & \scriptsize \textbf{\evared{-0.4}} & \scriptsize \textbf{\evagreen{+1.4}} & \scriptsize \textbf{\evagreen{+0.5}}\\
    \end{tabular}
\caption{\textbf{Increasing resolution.} We report zero-shot performance on ImageNet variants and ObjectNet.}
\label{tab: ablation_resolution}
\end{table}

\paragraph{Image Resolution.} In \tblref{tab: ablation_resolution}, we investigate the impact of larger image resolutions on zero-shot performance. Notably, there is an average top-1 accuracy gain of \textbf{\evagreen{+0.9}} when the resolution increases from 224\suptext{2} to 448\suptext{2} for \evaclipeight. Similarly, an increase from 224\suptext{2} to 336\suptext{2} results in a gain of \textbf{\evagreen{+0.5}}, even when trained with low global batch sizes of 24k for \evaclipeight+ and 23k for {\evaclipx}+.

\section{Conclusion}

We present \evaclipx, the currently largest and most performant open-sourced CLIP model with 18-billion parameters.
We show that following \evaone's weak-to-strong vision scaling principle, we can further scale up CLIP models to a new record and advance SOTA on multiple prevalent benchmarks across image, video and 3D domains.
Importantly, we demonstrate that scaling up the size of \evaclip models consistently boosts performance with no sign of saturation, shedding light on future vision model scaling.

{
\fontsize{8.2pt}{9.84pt}\selectfont
\bibliographystyle{ieee_fullname}
\bibliography{evaclip}

\begin{thebibliography}{10}\itemsep=-1pt

\bibitem{laioncoco}
Laion coco: 600m synthetic captions from laion2b-en.
\newblock \url{https://laion.ai/blog/laion-coco/}.

\bibitem{clipbigg}
Reaching 80 zero-shot accuracy with openclip: Vit-g/14 trained on laion-2b.
\newblock \url{https://laion.ai/blog/giant-openclip/}.

\bibitem{alayrac2022flamingo}
Jean-Baptiste Alayrac, Jeff Donahue, Pauline Luc, Antoine Miech, Iain Barr, Yana Hasson, Karel Lenc, Arthur Mensch, Katie Millican, Malcolm Reynolds, et~al.
\newblock Flamingo: a visual language model for few-shot learning.
\newblock {\em arXiv preprint arXiv:2204.14198}, 2022.

\bibitem{layernorm}
Jimmy~Lei Ba, Jamie~Ryan Kiros, and Geoffrey~E. Hinton.
\newblock Layer normalization.
\newblock {\em arXiv preprint arXiv:1607.06450}, 2016.

\bibitem{Qwen-VL}
Jinze Bai, Shuai Bai, Shusheng Yang, Shijie Wang, Sinan Tan, Peng Wang, Junyang Lin, Chang Zhou, and Jingren Zhou.
\newblock Qwen-vl: A versatile vision-language model for understanding, localization, text reading, and beyond.
\newblock {\em arXiv preprint arXiv:2308.12966}, 2023.

\bibitem{bain2021frozen}
Max Bain, Arsha Nagrani, G{\"u}l Varol, and Andrew Zisserman.
\newblock Frozen in time: A joint video and image encoder for end-to-end retrieval.
\newblock In {\em Proceedings of the IEEE/CVF International Conference on Computer Vision}, pages 1728--1738, 2021.

\bibitem{bao2021beit}
Hangbo Bao, Li Dong, and Furu Wei.
\newblock Beit: Bert pre-training of image transformers.
\newblock {\em arXiv preprint arXiv:2106.08254}, 2021.

\bibitem{objectnet}
Andrei Barbu, David Mayo, Julian Alverio, William Luo, Christopher Wang, Dan Gutfreund, Josh Tenenbaum, and Boris Katz.
\newblock Objectnet: A large-scale bias-controlled dataset for pushing the limits of object recognition models.
\newblock In {\em NeurIPS}, 2019.

\bibitem{berg2014birdsnap}
Thomas Berg, Jiongxin Liu, Seung Woo~Lee, Michelle~L Alexander, David~W Jacobs, and Peter~N Belhumeur.
\newblock Birdsnap: Large-scale fine-grained visual categorization of birds.
\newblock In {\em CVPR}, 2014.

\bibitem{bossard2014food}
Lukas Bossard, Matthieu Guillaumin, and Luc Van~Gool.
\newblock Food-101--mining discriminative components with random forests.
\newblock In {\em ECCV}, 2014.

\bibitem{gpt3}
Tom~B. Brown, Benjamin Mann, Nick Ryder, Melanie Subbiah, Jared Kaplan, Prafulla Dhariwal, Arvind Neelakantan, Pranav Shyam, Girish Sastry, Amanda Askell, Sandhini Agarwal, Ariel Herbert-Voss, Gretchen Krueger, Tom Henighan, Rewon Child, Aditya Ramesh, Daniel~M. Ziegler, Jeffrey Wu, Clemens Winter, Christopher Hesse, Mark Chen, Eric Sigler, Mateusz Litwin, Scott Gray, Benjamin Chess, Jack Clark, Christopher Berner, Sam McCandlish, Alec Radford, Ilya Sutskever, and Dario Amodei.
\newblock Language models are few-shot learners.
\newblock {\em arXiv preprint arXiv:2005.14165}, 2020.

\bibitem{kakaobrain2022coyo-700m}
Minwoo Byeon, Beomhee Park, Haecheon Kim, Sungjun Lee, Woonhyuk Baek, and Saehoon Kim.
\newblock Coyo-700m: Image-text pair dataset.
\newblock \url{https://github.com/kakaobrain/coyo-dataset}, 2022.

\bibitem{k600}
Joao Carreira, Eric Noland, Andras Banki-Horvath, Chloe Hillier, and Andrew Zisserman.
\newblock A short note about kinetics-600.
\newblock {\em arXiv preprint arXiv:1808.01340}, 2018.

\bibitem{k700}
Joao Carreira, Eric Noland, Chloe Hillier, and Andrew Zisserman.
\newblock A short note on the kinetics-700 human action dataset.
\newblock {\em arXiv preprint arXiv:1907.06987}, 2019.

\bibitem{carreira2017quo}
Joao Carreira and Andrew Zisserman.
\newblock Quo vadis, action recognition? a new model and the kinetics dataset.
\newblock In {\em CVPR}, 2017.

\bibitem{gradcheckpointing}
Tianqi Chen, Bing Xu, Chiyuan Zhang, and Carlos Guestrin.
\newblock Training deep nets with sublinear memory cost, 2016.

\bibitem{chen2023internvl}
Zhe Chen, Jiannan Wu, Wenhai Wang, Weijie Su, Guo Chen, Sen Xing, Muyan Zhong, Qinglong Zhang, Xizhou Zhu, Lewei Lu, Bin Li, Ping Luo, Tong Lu, Yu Qiao, and Jifeng Dai.
\newblock Internvl: Scaling up vision foundation models and aligning for generic visual-linguistic tasks.
\newblock {\em arXiv preprint arXiv:2312.14238}, 2023.

\bibitem{cheng2017remote}
Gong Cheng, Junwei Han, and Xiaoqiang Lu.
\newblock Remote sensing image scene classification: Benchmark and state of the art.
\newblock {\em Proceedings of the IEEE}, 2017.

\bibitem{openclip2022scalelaws}
Mehdi Cherti, Romain Beaumont, Ross Wightman, Mitchell Wortsman, Gabriel Ilharco, Cade Gordon, Christoph Schuhmann, Ludwig Schmidt, and Jenia Jitsev.
\newblock Reproducible scaling laws for contrastive language-image learning, 2022.

\bibitem{chowdhery2022palm}
Aakanksha Chowdhery, Sharan Narang, Jacob Devlin, Maarten Bosma, Gaurav Mishra, Adam Roberts, Paul Barham, Hyung~Won Chung, Charles Sutton, Sebastian Gehrmann, Parker Schuh, Kensen Shi, Sasha Tsvyashchenko, Joshua Maynez, Abhishek Rao, Parker Barnes, Yi Tay, Noam Shazeer, Vinodkumar Prabhakaran, Emily Reif, Nan Du, Ben Hutchinson, Reiner Pope, James Bradbury, Jacob Austin, Michael Isard, Guy Gur-Ari, Pengcheng Yin, Toju Duke, Anselm Levskaya, Sanjay Ghemawat, Sunipa Dev, Henryk Michalewski, Xavier Garcia, Vedant Misra, Kevin Robinson, Liam Fedus, Denny Zhou, Daphne Ippolito, David Luan, Hyeontaek Lim, Barret Zoph, Alexander Spiridonov, Ryan Sepassi, David Dohan, Shivani Agrawal, Mark Omernick, Andrew~M. Dai, Thanumalayan~Sankaranarayana Pillai, Marie Pellat, Aitor Lewkowycz, Erica Moreira, Rewon Child, Oleksandr Polozov, Katherine Lee, Zongwei Zhou, Xuezhi Wang, Brennan Saeta, Mark Diaz, Orhan Firat, Michele Catasta, Jason Wei, Kathy Meier-Hellstern, Douglas Eck, Jeff Dean, Slav Petrov, and Noah Fiedel.
\newblock Palm: Scaling language modeling with pathways.
\newblock {\em arXiv preprint arXiv:2204.02311}, 2022.

\bibitem{cimpoi14describing}
M. Cimpoi, S. Maji, I. Kokkinos, S. Mohamed, , and A. Vedaldi.
\newblock Describing textures in the wild.
\newblock In {\em CVPR}, 2014.

\bibitem{clark2020electra}
Kevin Clark, Minh-Thang Luong, Quoc~V Le, and Christopher~D Manning.
\newblock {ELECTRA}: Pre-training text encoders as discriminators rather than generators.
\newblock {\em arXiv preprint arXiv:2003.10555}, 2020.

\bibitem{coates2011analysis}
Adam Coates, Andrew Ng, and Honglak Lee.
\newblock An analysis of single-layer networks in unsupervised feature learning.
\newblock In {\em AISTAT}, 2011.

\bibitem{flashattention}
Tri Dao, Daniel~Y. Fu, Stefano Ermon, Atri Rudra, and Christopher Ré.
\newblock Flashattention: Fast and memory-efficient exact attention with io-awareness, 2022.

\bibitem{deitke2023objaverse}
Matt Deitke, Dustin Schwenk, Jordi Salvador, Luca Weihs, Oscar Michel, Eli VanderBilt, Ludwig Schmidt, Kiana Ehsani, Aniruddha Kembhavi, and Ali Farhadi.
\newblock Objaverse: A universe of annotated 3d objects.
\newblock In {\em Proceedings of the IEEE/CVF Conference on Computer Vision and Pattern Recognition}, pages 13142--13153, 2023.

\bibitem{deng2009imagenet}
Jia Deng, Wei Dong, Richard Socher, Li-Jia Li, Kai Li, and Li Fei-Fei.
\newblock Imagenet: A large-scale hierarchical image database.
\newblock In {\em CVPR}, 2009.

\bibitem{pascal-voc-2007}
M. Everingham, L. Van~Gool, C.~K.~I. Williams, J. Winn, and A. Zisserman.
\newblock "the {PASCAL} {V}isual {O}bject {C}lasses {C}hallenge 2007 {(VOC2007)} {R}esults.
\newblock "http://www.pascal-network.org/challenges/VOC/voc2007/workshop/index.html", 2007.

\bibitem{dfn5b}
Alex Fang, Albin~Madappally Jose, Amit Jain, Ludwig Schmidt, Alexander Toshev, and Vaishaal Shankar.
\newblock Data filtering networks.
\newblock {\em arXiv preprint arXiv:2309.17425}, 2023.

\bibitem{EVA02}
Yuxin Fang, Quan Sun, Xinggang Wang, Tiejun Huang, Xinlong Wang, and Yue Cao.
\newblock Eva-02: A visual representation for neon genesis.
\newblock {\em arXiv preprint arXiv:2303.11331}, 2023.

\bibitem{eva}
Yuxin Fang, Wen Wang, Binhui Xie, Quan Sun, Ledell Wu, Xinggang Wang, Tiejun Huang, Xinlong Wang, and Yue Cao.
\newblock Eva: Exploring the limits of masked visual representation learning at scale.
\newblock {\em arXiv preprint arXiv:2211.07636}, 2022.

\bibitem{fei2004learning}
Li Fei-Fei, Rob Fergus, and Pietro Perona.
\newblock Learning generative visual models from few training examples: An incremental bayesian approach tested on 101 object categories.
\newblock In {\em CVPRW}, 2004.

\bibitem{gadre2023datacomp}
Samir~Yitzhak Gadre, Gabriel Ilharco, Alex Fang, Jonathan Hayase, Georgios Smyrnis, Thao Nguyen, Ryan Marten, Mitchell Wortsman, Dhruba Ghosh, Jieyu Zhang, Eyal Orgad, Rahim Entezari, Giannis Daras, Sarah Pratt, Vivek Ramanujan, Yonatan Bitton, Kalyani Marathe, Stephen Mussmann, Richard Vencu, Mehdi Cherti, Ranjay Krishna, Pang~Wei Koh, Olga Saukh, Alexander Ratner, Shuran Song, Hannaneh Hajishirzi, Ali Farhadi, Romain Beaumont, Sewoong Oh, Alex Dimakis, Jenia Jitsev, Yair Carmon, Vaishaal Shankar, and Ludwig Schmidt.
\newblock Datacomp: In search of the next generation of multimodal datasets.
\newblock {\em arXiv preprint arXiv:2304.14108}, 2023.

\bibitem{goodfellow2013challenges}
Ian~J Goodfellow, Dumitru Erhan, Pierre~Luc Carrier, Aaron Courville, Mehdi Mirza, Ben Hamner, Will Cukierski, Yichuan Tang, David Thaler, Dong-Hyun Lee, et~al.
\newblock Challenges in representation learning: A report on three machine learning contests.
\newblock In {\em ICONIP}, 2013.

\bibitem{helber2019eurosat}
Patrick Helber, Benjamin Bischke, Andreas Dengel, and Damian Borth.
\newblock Eurosat: A novel dataset and deep learning benchmark for land use and land cover classification.
\newblock {\em IEEE J. Sel. Top. Appl. Earth Obs. Remote Sens.}, 2019.

\bibitem{inren}
Dan Hendrycks, Steven Basart, Norman Mu, Saurav Kadavath, Frank Wang, Evan Dorundo, Rahul Desai, Tyler Zhu, Samyak Parajuli, Mike Guo, et~al.
\newblock The many faces of robustness: A critical analysis of out-of-distribution generalization.
\newblock In {\em CVPR}, 2021.

\bibitem{inadv}
Dan Hendrycks, Kevin Zhao, Steven Basart, Jacob Steinhardt, and Dawn Song.
\newblock Natural adversarial examples.
\newblock In {\em CVPR}, 2021.

\bibitem{huang2016deep}
Gao Huang, Yu Sun, Zhuang Liu, Daniel Sedra, and Kilian~Q Weinberger.
\newblock Deep networks with stochastic depth.
\newblock In {\em ECCV}, 2016.

\bibitem{openclip}
Gabriel Ilharco, Mitchell Wortsman, Ross Wightman, Cade Gordon, Nicholas Carlini, Rohan Taori, Achal Dave, Vaishaal Shankar, Hongseok Namkoong, John Miller, Hannaneh Hajishirzi, Ali Farhadi, and Ludwig Schmidt.
\newblock Openclip.
\newblock \url{https://github.com/mlfoundations/open_clip}, 2021.

\bibitem{krause20133d}
Jonathan Krause, Michael Stark, Jia Deng, and Li Fei-Fei.
\newblock 3d object representations for fine-grained categorization.
\newblock In {\em ICCVW}, 2013.

\bibitem{cifar}
Alex Krizhevsky, Geoffrey Hinton, et~al.
\newblock Learning multiple layers of features from tiny images.
\newblock 2009.

\bibitem{lecun1998gradient}
Yann LeCun, L{\'e}on Bottou, Yoshua Bengio, and Patrick Haffner.
\newblock Gradient-based learning applied to document recognition.
\newblock {\em Proceedings of the IEEE}, 1998.

\bibitem{blip2}
Junnan Li, Dongxu Li, Silvio Savarese, and Steven Hoi.
\newblock Blip-2: Bootstrapping language-image pre-training with frozen image encoders and large language models.
\newblock {\em arXiv preprint arXiv:2301.12597}, 2023.

\bibitem{li2023clipav2}
Xianhang Li, Zeyu Wang, and Cihang Xie.
\newblock Clipa-v2: Scaling clip training with 81.1
\newblock {\em arXiv preprint arXiv:2306.15658}, 2023.

\bibitem{flip}
Yanghao Li, Haoqi Fan, Ronghang Hu, Christoph Feichtenhofer, and Kaiming He.
\newblock Scaling language-image pre-training via masking, 2022.

\bibitem{lin2014coco}
Tsung-Yi Lin, Michael Maire, Serge Belongie, James Hays, Pietro Perona, Deva Ramanan, Piotr Doll{\'a}r, and C~Lawrence Zitnick.
\newblock Microsoft coco: Common objects in context.
\newblock In {\em European conference on computer vision}, pages 740--755. Springer, 2014.

\bibitem{liu2023llava}
Haotian Liu, Chunyuan Li, Qingyang Wu, and Yong~Jae Lee.
\newblock Visual instruction tuning.
\newblock {\em arXiv preprint arXiv:2304.08485}, 2023.

\bibitem{maji2013fine}
Subhransu Maji, Esa Rahtu, Juho Kannala, Matthew Blaschko, and Andrea Vedaldi.
\newblock Fine-grained visual classification of aircraft.
\newblock {\em arXiv preprint arXiv:1306.5151}, 2013.

\bibitem{videocc}
Arsha Nagrani, Paul~Hongsuck Seo, Bryan Seybold, Anja Hauth, Santiago Manen, Chen Sun, and Cordelia Schmid.
\newblock Learning audio-video modalities from image captions, 2022.

\bibitem{nilsback2008automated}
Maria-Elena Nilsback and Andrew Zisserman.
\newblock Automated flower classification over a large number of classes.
\newblock In {\em ICVGIP}, 2008.

\bibitem{pan2023tap}
Ting Pan, Lulu Tang, Xinlong Wang, and Shiguang Shan.
\newblock Tokenize anything via prompting.
\newblock {\em arXiv preprint arXiv:2312.09128}, 2023.

\bibitem{parkhi12a}
Omkar~M. Parkhi, Andrea Vedaldi, Andrew Zisserman, and C.~V. Jawahar.
\newblock Cats and dogs.
\newblock In {\em CVPR}, 2012.

\bibitem{basic}
Hieu Pham, Zihang Dai, Golnaz Ghiasi, Hanxiao Liu, Adams~Wei Yu, Minh-Thang Luong, Mingxing Tan, and Quoc~V Le.
\newblock Combined scaling for zero-shot transfer learning.
\newblock {\em arXiv preprint arXiv:2111.10050}, 2021.

\bibitem{clip}
Alec Radford, Jong~Wook Kim, Chris Hallacy, Aditya Ramesh, Gabriel Goh, Sandhini Agarwal, Girish Sastry, Amanda Askell, Pamela Mishkin, Jack Clark, et~al.
\newblock Learning transferable visual models from natural language supervision.
\newblock In {\em ICML}, 2021.

\bibitem{t5}
Colin Raffel, Noam Shazeer, Adam Roberts, Katherine Lee, Sharan Narang, Michael Matena, Yanqi Zhou, Wei Li, Peter~J Liu, et~al.
\newblock Exploring the limits of transfer learning with a unified text-to-text transformer.
\newblock {\em JMLR}, 2020.

\bibitem{rajbhandari2020zero}
Samyam Rajbhandari, Jeff Rasley, Olatunji Ruwase, and Yuxiong He.
\newblock Zero: Memory optimizations toward training trillion parameter models.
\newblock In {\em SC20}, 2020.

\bibitem{rasley2020deepspeed}
Jeff Rasley, Samyam Rajbhandari, Olatunji Ruwase, and Yuxiong He.
\newblock Deepspeed: System optimizations enable training deep learning models with over 100 billion parameters.
\newblock In {\em KDD}, 2020.

\bibitem{recht2019imagenetv2}
Benjamin Recht, Rebecca Roelofs, Ludwig Schmidt, and Vaishaal Shankar.
\newblock Do imagenet classifiers generalize to imagenet?, 2019.

\bibitem{laion5b}
Christoph Schuhmann, Romain Beaumont, Richard Vencu, Cade Gordon, Ross Wightman, Mehdi Cherti, Theo Coombes, Aarush Katta, Clayton Mullis, Mitchell Wortsman, et~al.
\newblock Laion-5b: An open large-scale dataset for training next generation image-text models.
\newblock {\em arXiv preprint arXiv:2210.08402}, 2022.

\bibitem{laion400m}
Christoph Schuhmann, Richard Vencu, Romain Beaumont, Robert Kaczmarczyk, Clayton Mullis, Aarush Katta, Theo Coombes, Jenia Jitsev, and Aran Komatsuzaki.
\newblock Laion-400m: Open dataset of clip-filtered 400 million image-text pairs.
\newblock {\em arXiv preprint arXiv:2111.02114}, 2021.

\bibitem{ucf101}
Khurram Soomro, Amir~Roshan Zamir, and Mubarak Shah.
\newblock Ucf101: A dataset of 101 human actions classes from videos in the wild.
\newblock {\em arXiv preprint arXiv:1212.0402}, 2012.

\bibitem{stallkamp2012man}
Johannes Stallkamp, Marc Schlipsing, Jan Salmen, and Christian Igel.
\newblock Man vs. computer: Benchmarking machine learning algorithms for traffic sign recognition.
\newblock {\em Neural networks}, 2012.

\bibitem{emu2}
Quan Sun, Yufeng Cui, Xiaosong Zhang, Fan Zhang, Qiying Yu, Zhengxiong Luo, Yueze Wang, Yongming Rao, Jingjing Liu, Tiejun Huang, and Xinlong Wang.
\newblock Generative multimodal models are in-context learners.
\newblock {\em arXiv preprint arXiv:2312.13286}, 2023.

\bibitem{sun2023evaclip}
Quan Sun, Yuxin Fang, Ledell Wu, Xinlong Wang, and Yue Cao.
\newblock Eva-clip: Improved training techniques for clip at scale.
\newblock {\em arXiv preprint arXiv:2303.15389}, 2023.

\bibitem{emu1}
Quan Sun, Qiying Yu, Yufeng Cui, Fan Zhang, Xiaosong Zhang, Yueze Wang, Hongcheng Gao, Jingjing Liu, Tiejun Huang, and Xinlong Wang.
\newblock Generative pretraining in multimodality.
\newblock {\em arXiv preprint arXiv:2307.05222}, 2023.

\bibitem{touvron2023llama}
Hugo Touvron, Thibaut Lavril, Gautier Izacard, Xavier Martinet, Marie-Anne Lachaux, Timothée Lacroix, Baptiste Rozière, Naman Goyal, Eric Hambro, Faisal Azhar, Aurelien Rodriguez, Armand Joulin, Edouard Grave, and Guillaume Lample.
\newblock Llama: Open and efficient foundation language models.
\newblock {\em arXiv preprint arXiv:2302.13971}, 2023.

\bibitem{uy2019revisiting}
Mikaela~Angelina Uy, Quang-Hieu Pham, Binh-Son Hua, Thanh Nguyen, and Sai-Kit Yeung.
\newblock Revisiting point cloud classification: A new benchmark dataset and classification model on real-world data.
\newblock In {\em Proceedings of the IEEE/CVF international conference on computer vision}, pages 1588--1597, 2019.

\bibitem{veeling2018rotation}
Bastiaan~S Veeling, Jasper Linmans, Jim Winkens, Taco Cohen, and Max Welling.
\newblock Rotation equivariant cnns for digital pathology.
\newblock In {\em MICCAI}, 2018.

\bibitem{inske}
Haohan Wang, Songwei Ge, Zachary Lipton, and Eric~P Xing.
\newblock Learning robust global representations by penalizing local predictive power.
\newblock {\em NeurIPS}, 2019.

\bibitem{wang2023cogvlm}
Weihan Wang, Qingsong Lv, Wenmeng Yu, Wenyi Hong, Ji Qi, Yan Wang, Junhui Ji, Zhuoyi Yang, Lei Zhao, Xixuan Song, et~al.
\newblock Cogvlm: Visual expert for pretrained language models.
\newblock {\em arXiv preprint arXiv:2311.03079}, 2023.

\bibitem{wang2024internvid}
Yi Wang, Yinan He, Yizhuo Li, Kunchang Li, Jiashuo Yu, Xin Ma, Xinhao Li, Guo Chen, Xinyuan Chen, Yaohui Wang, Conghui He, Ping Luo, Ziwei Liu, Yali Wang, Limin Wang, and Yu Qiao.
\newblock Internvid: A large-scale video-text dataset for multimodal understanding and generation, 2024.

\bibitem{wu20153d}
Zhirong Wu, Shuran Song, Aditya Khosla, Fisher Yu, Linguang Zhang, Xiaoou Tang, and Jianxiong Xiao.
\newblock 3d shapenets: A deep representation for volumetric shapes.
\newblock In {\em Proceedings of the IEEE conference on computer vision and pattern recognition}, pages 1912--1920, 2015.

\bibitem{xiao2010sun}
Jianxiong Xiao, James Hays, Krista~A Ehinger, Aude Oliva, and Antonio Torralba.
\newblock Sun database: Large-scale scene recognition from abbey to zoo.
\newblock In {\em CVPR}, 2010.

\bibitem{lamb}
Yang You, Jing Li, Sashank Reddi, Jonathan Hseu, Sanjiv Kumar, Srinadh Bhojanapalli, Xiaodan Song, James Demmel, Kurt Keutzer, and Cho-Jui Hsieh.
\newblock Large batch optimization for deep learning: Training bert in 76 minutes, 2019.

\bibitem{flickr30K}
Peter Young, Alice Lai, Micah Hodosh, and Julia Hockenmaier.
\newblock From image descriptions to visual denotations: New similarity metrics for semantic inference over event descriptions.
\newblock {\em TACL}, 2014.

\bibitem{siglip}
Xiaohua Zhai, Basil Mustafa, Alexander Kolesnikov, and Lucas Beyer.
\newblock Sigmoid loss for language image pre-training.
\newblock {\em arXiv preprint arXiv:2303.15343}, 2023.

\bibitem{rmsnorm}
Biao Zhang and Rico Sennrich.
\newblock Root mean square layer normalization.
\newblock {\em arXiv preprint arXiv:1910.07467}, 2019.

\bibitem{uni3d}
Junsheng Zhou, Jinsheng Wang, Baorui Ma, Yu-Shen Liu, Tiejun Huang, and Xinlong Wang.
\newblock Uni3d: Exploring unified 3d representation at scale.
\newblock {\em arXiv preprint arXiv:2310.06773}, 2023.

\bibitem{zhu2023minigpt4}
Deyao Zhu, Jun Chen, Xiaoqian Shen, Xiang Li, and Mohamed Elhoseiny.
\newblock Minigpt-4: Enhancing vision-language understanding with advanced large language models.
\newblock {\em arXiv preprint arXiv:2304.10592}, 2023.

\end{thebibliography}
}

\clearpage
\maketitlesupplementary

\begin{table}[!h]
\vspace{.5em}
\centering
\tablestyle{1.5pt}{1.2}
\begin{tabular}{lllr}
\toprule
    \scriptsize Dataset & \scriptsize  Classes & \scriptsize  Test size & \scriptsize  Evaluation Metric \\
\midrule
\scriptsize ImageNet-1K~\cite{deng2009imagenet} & \scriptsize 1000 & \scriptsize 50,000 & \scriptsize accuracy \\
\scriptsize ImageNet-V2~\cite{recht2019imagenetv2} & \scriptsize 1000 & \scriptsize 10,000 & \scriptsize accuracy \\
\scriptsize ImageNet-Adversarial~\cite{inadv} & \scriptsize 1000 & \scriptsize 7,500 & \scriptsize accuracy \\
\scriptsize ImageNet-R(endition)~\cite{inren} & \scriptsize 1000 & \scriptsize 30,000 & \scriptsize accuracy \\
\scriptsize ImageNet-Sketch~\cite{inske} & \scriptsize 1000 & \scriptsize 50,899 & \scriptsize accuracy \\
\scriptsize ObjectNet~\cite{objectnet} & \scriptsize 1000 & \scriptsize 50,273 & \scriptsize accuracy \\

\scriptsize CIFAR-10~\cite{cifar} & \scriptsize 10 & \scriptsize 10,000 & \scriptsize accuracy \\
\scriptsize CIFAR-100~\cite{cifar} & \scriptsize 100 & \scriptsize 10,000 & \scriptsize accuracy \\
\scriptsize MNIST~\cite{lecun1998gradient} & \scriptsize 10 & \scriptsize 10,000 & \scriptsize accuracy \\
\scriptsize Caltech101~\cite{fei2004learning} & \scriptsize 101 & \scriptsize 9144 & \scriptsize accuracy \\
\scriptsize SUN397~\cite{xiao2010sun} & \scriptsize 397 & \scriptsize 108,754 & \scriptsize accuracy \\
\scriptsize FGVC Aircraft~\cite{maji2013fine} & \scriptsize 100 & \scriptsize 3,333 & \scriptsize accuracy \\
\scriptsize Country-211~\cite{clip} & \scriptsize 211 & \scriptsize 21,100 & \scriptsize accuracy \\
\scriptsize Stanford Cars~\cite{krause20133d} & \scriptsize 196 & \scriptsize 8,041 & \scriptsize accuracy \\
\scriptsize Birdsnap~\cite{berg2014birdsnap} & \scriptsize 500 & \scriptsize 2,195 & \scriptsize accuracy \\
\scriptsize Describable \scriptsize Textures~\cite{cimpoi14describing} & \scriptsize 47 & \scriptsize 1,880 & \scriptsize accuracy \\
\scriptsize EuroSAT\cite{helber2019eurosat} & \scriptsize 10 & \scriptsize 27,000 & \scriptsize accuracy \\
\scriptsize Facial Emotion Recognition 2013~\cite{goodfellow2013challenges} & \scriptsize 8 & \scriptsize 3,574 & \scriptsize accuracy \\
\scriptsize Oxford Flowers 102~\cite{nilsback2008automated} & \scriptsize 102 & \scriptsize 6,149 & \scriptsize accuracy \\
\scriptsize Food-101~\cite{bossard2014food} & \scriptsize 102 & \scriptsize 25,250 & \scriptsize accuracy \\
\scriptsize GTSRB~\cite{stallkamp2012man} & \scriptsize 43 & \scriptsize 12,630 & \scriptsize accuracy \\
\scriptsize PatchCamelyon~\cite{veeling2018rotation} & \scriptsize 2 & \scriptsize 32,768 & \scriptsize accuracy \\
\scriptsize Oxford-IIIT Pets~\cite{parkhi12a} & \scriptsize 37 & \scriptsize 3,669 & \scriptsize accuracy \\
\scriptsize Rendered SST2~\cite{clip} & \scriptsize 2 & \scriptsize 1,821 & \scriptsize accuracy \\
\scriptsize RESISC45~\cite{cheng2017remote} & \scriptsize 45 & \scriptsize 31,500 & \scriptsize accuracy \\
\scriptsize STL-10~\cite{coates2011analysis} & \scriptsize 10 & \scriptsize 8000 & \scriptsize accuracy \\
\scriptsize Pascal VOC 2007 Classification~\cite{pascal-voc-2007} & \scriptsize 20 & \scriptsize 4,952 & \scriptsize accuracy \\

\midrule
\scriptsize UC-F101~\cite{ucf101} & \scriptsize 101 & \scriptsize 11,213 & \scriptsize accuracy \\
\scriptsize Kinetics-400~\cite{carreira2017quo} & \scriptsize 400 & \scriptsize 19,240 & \scriptsize mean(top1, top5) \\
\scriptsize Kinetics-600~\cite{k600} & \scriptsize 600 & \scriptsize 29,788 & \scriptsize mean(top1, top5) \\
\scriptsize Kinetics-700~\cite{k700} & \scriptsize 700 & \scriptsize 33,966 & \scriptsize mean(top1, top5) \\
\midrule
\scriptsize Flickr30K~\cite{flickr30K} & \scriptsize - & \scriptsize 1000 & \scriptsize recall \\
\scriptsize COCO~\cite{lin2014coco} & \scriptsize - & \scriptsize 5000 & \scriptsize recall \\
\bottomrule
\end{tabular}
\caption{Datasets used to evaluate \evaclip models.}
\label{tab: eval_dataset_table}
\end{table}


\begin{table}[h]
\vspace{-.5em}
\centering
\tablestyle{6pt}{1.2}
\scriptsize
\begin{tabular}{l|c}
config & \evaclip-\{8B, 8B+\} \\
\shline

image enc. weight init. & \evaone-8B / \evaclipeight \\
text enc. weight init. & \evaTwoclip-E/14+ / \evaclipeight \\

image-text data & Merged-2B \\

image enc. peak learning rate &  4e-4 / 2e-4 \\
image enc. layer-wise lr decay~\cite{clark2020electra, bao2021beit} & 0.9 / 0.85 \\
text enc. peak learning rate &  4e-5 / 2e-5 \\
text enc. layer-wise lr decay~\cite{clark2020electra, bao2021beit} & 0.75 \\

learning rate schedule & cosine decay \\

optimizer & LAMB~\cite{lamb} \\
optimizer hyper-parameters & $\beta_1$, $\beta_2$, $\epsilon$ = 0.9, 0.95, 1e-6 \\
weight decay & 0 \\

input resolution & 224\suptext{2} / 448\suptext{2} \\
patch size & 14\suptext{2} \\

batch size & 178k / 24k \\
samples seen & 9B / 800M \\

drop image patch~\cite{flip} & 0.5 / 0.0 \\

drop path~\cite{huang2016deep} & 0.0 \\
random resized crop & (0.9, 1) \\

numerical precision & $\mathtt{DeepSpeed}$ $\mathtt{bf16}$~\cite{rasley2020deepspeed} \\
ZeRO optimizer~\cite{rajbhandari2020zero} & stage 3 \\

\end{tabular}
\caption{\evaclipeight and {\evaclipeight}+ training settings.}
\label{tab: clip cfg 1}
\end{table}

\begin{table}[h]
\centering
\tablestyle{6pt}{1.2}
\scriptsize
\begin{tabular}{l|c}
config & \evaclip-\{18B,18B+\} \\
\shline

image enc. weight init. & \evaone-18B / \evaclipx \\
text enc. weight init. & \evaTwoclip-E/14+ / \evaclipx \\

image-text data & Merged-2B+ \\

image enc. peak learning rate &  4e-4 / 2e-4 \\
image enc. layer-wise lr decay~\cite{clark2020electra, bao2021beit} & 0.9 / 0.85 \\
text enc. peak learning rate &  4e-5 / 2e-5 \\
text enc. layer-wise lr decay~\cite{clark2020electra, bao2021beit} & 0.75 \\

learning rate schedule & cosine decay \\

optimizer & LAMB~\cite{lamb} \\
optimizer hyper-parameters & $\beta_1$, $\beta_2$, $\epsilon$ = 0.9, 0.95, 1e-6 \\
weight decay & 0 \\

input resolution & 224\suptext{2} / 336\suptext{2} \\
patch size & 14\suptext{2} \\

batch size & 108k / 23k \\
samples seen & 6B / 400M \\

drop image patch~\cite{flip} & 0.5 / 0.0 \\
drop path~\cite{huang2016deep} & 0.0 \\
random resized crop & (0.9, 1) \\

numerical precision & $\mathtt{DeepSpeed}$ $\mathtt{bf16}$~\cite{rasley2020deepspeed} \\
ZeRO optimizer~\cite{rajbhandari2020zero} & stage 3 \\

\end{tabular}
\caption{\evaclipx and {\evaclipx}+ training settings.}
\label{tab: clip cfg 2}
\end{table}


\begin{table*}[h]
\vspace{-.5em}
\centering
\subfloat[
\textbf{Impact of image transformations on zero-shot image classification performance.} Different transformations can significantly influence zero-shot image classification performance, particularly for ObjectNet~\cite{objectnet}. \evaclipx shows robustness with the same average top-1 accuracy across different image transformations.
\label{tab: ablate_transforms_img_cls}
]{
\centering
\begin{minipage}{1\linewidth}{\begin{center}
    \tablestyle{1.4pt}{1.2}
    \begin{tabular}{l|cccccc|ccccccccccccccccccccc|c}
        \scriptsize method &
        \rotatebox[origin=l]{90}{\scriptsize{ImageNet-1K~\cite{deng2009imagenet}}} &
        \rotatebox[origin=l]{90}{\scriptsize{ImageNet-V2~\cite{recht2019imagenetv2}}} &
        \rotatebox[origin=l]{90}{\scriptsize{ImageNet-Adv.~\cite{inadv}}} &
        \rotatebox[origin=l]{90}{\scriptsize{ImageNet-Ren.~\cite{inren}}} &
        \rotatebox[origin=l]{90}{\scriptsize{ImageNet-Ske.~\cite{inske}}} &
        \rotatebox[origin=l]{90}{\scriptsize{ObjectNet~\cite{objectnet}}} &
        \rotatebox[origin=l]{90}{\scriptsize{CIFAR-10~\cite{cifar}}} &
        \rotatebox[origin=l]{90}{\scriptsize{CIFAR-100~\cite{cifar}}} & 
        \rotatebox[origin=l]{90}{\scriptsize{MNIST~\cite{lecun1998gradient}}} & 
        \rotatebox[origin=l]{90}{\scriptsize{Caltech101~\cite{fei2004learning}}} & 
        \rotatebox[origin=l]{90}{\scriptsize{SUN397~\cite{xiao2010sun}}} & 
        \rotatebox[origin=l]{90}{\scriptsize{FGVC Aircraft~\cite{maji2013fine}}} & 
        \rotatebox[origin=l]{90}{\scriptsize{Country-211~\cite{clip}}} & 
        \rotatebox[origin=l]{90}{\scriptsize{Stanford Cars~\cite{krause20133d}}} &
        \rotatebox[origin=l]{90}{\scriptsize{Birdsnap~\cite{berg2014birdsnap}}} & 
        \rotatebox[origin=l]{90}{\scriptsize{DTD~\cite{cimpoi14describing}}} & 
        \rotatebox[origin=l]{90}{\scriptsize{EuroSAT~\cite{helber2019eurosat}}} & 
        \rotatebox[origin=l]{90}{\scriptsize{FER2013~\cite{goodfellow2013challenges}}} & 
        \rotatebox[origin=l]{90}{\scriptsize{Flowers-102~\cite{nilsback2008automated}}} & 
        \rotatebox[origin=l]{90}{\scriptsize{Food-101~\cite{bossard2014food}}} & 
        \rotatebox[origin=l]{90}{\scriptsize{GTSRB~\cite{stallkamp2012man}}} & 
        \rotatebox[origin=l]{90}{\scriptsize{PCam~\cite{veeling2018rotation}}} & 
        \rotatebox[origin=l]{90}{\scriptsize{Pets~\cite{parkhi12a}}} & 
        \rotatebox[origin=l]{90}{\scriptsize{Rendered SST2~\cite{clip}}} & 
        \rotatebox[origin=l]{90}{\scriptsize{RESISC45~\cite{cheng2017remote}}} & 
        \rotatebox[origin=l]{90}{\scriptsize{STL-10~\cite{coates2011analysis}}} & 
        \rotatebox[origin=l]{90}{\scriptsize{VOC2007~\cite{pascal-voc-2007}}} &
        \rotatebox[origin=l]{90}{\ph{.}\textbf{avg. top-1 acc.}}
        \\
        \shline

        \scriptsize DFN5B-CLIP-H/14+* & \scriptsize 84.0  & \scriptsize 77.8 & \scriptsize \textbf{79.6} & \scriptsize 92.9 & \scriptsize 72.4 & \scriptsize \textbf{79.6} & \scriptsize 98.8 & \scriptsize 90.5 & \scriptsize 83.6 & \scriptsize 88.7 & \scriptsize 77.0 & \scriptsize 64.9 & \scriptsize 36.1 & \scriptsize 95.7 & \scriptsize \textbf{80.5} & \scriptsize 70.9 & \scriptsize 61.1 & \scriptsize 56.1 & \scriptsize \textbf{91.6} & \scriptsize 96.1 & \scriptsize 67.8 & \scriptsize 69.6 & \scriptsize 96.7 & \scriptsize 55.5 & \scriptsize 75.9 & \scriptsize 99.1 & \scriptsize 78.2 & \scriptsize 78.5 \\

        \scriptsize DFN5B-CLIP-H/14+$\dag$ & \scriptsize \textbf{84.3} & \scriptsize \textbf{78.3} & \scriptsize 79.3 & \scriptsize \textbf{93.6} & \scriptsize \textbf{73.3} & \scriptsize 73.5 & \scriptsize 98.8 & \scriptsize 90.5 & \scriptsize 83.6 & \scriptsize \textbf{88.9} & \scriptsize \textbf{77.4} & \scriptsize \textbf{72.5} & \scriptsize \textbf{37.9} & \scriptsize \textbf{96.0} & \scriptsize 80.3 & \scriptsize 70.9 & \scriptsize 61.1 & \scriptsize 56.1 & \scriptsize 91.4 & \scriptsize \textbf{96.2} & \scriptsize \textbf{67.9} & \scriptsize 69.6 & \scriptsize \textbf{96.8} & \scriptsize 55.5 & \scriptsize 75.9 & \scriptsize 99.1 & \scriptsize \textbf{81.9} & \scriptsize \textbf{78.9}  \\

        \midrule

        \rgray
        \scriptsize \evaclipx * & \scriptsize 83.7 & \scriptsize \textbf{77.9} & \scriptsize \textbf{87.3} & \scriptsize 95.6 & \scriptsize 74.4 & \scriptsize \textbf{82.2} & \scriptsize 99.4 & \scriptsize 93.8 & \scriptsize 83.0 & \scriptsize 89.4 & \scriptsize 77.5 & \scriptsize 58.4 & \scriptsize 41.8 & \scriptsize 94.9 & \scriptsize \textbf{79.9} & \scriptsize 71.9 & \scriptsize 79.8 & \scriptsize 59.3 & \scriptsize 85.9 & \scriptsize \textbf{95.8} & \scriptsize \textbf{72.4} & \scriptsize 65.2 & \scriptsize 96.0 & \scriptsize 67.5 & \scriptsize 76.8 & \scriptsize 99.6 & \scriptsize 82.4 & \scriptsize \textbf{80.4} \\

        \rgray
        \scriptsize \evaclipx $\dag$ & \scriptsize \textbf{83.8} & \scriptsize 77.7 & \scriptsize 86.2 & \scriptsize \textbf{95.7} & \scriptsize \textbf{74.7} & \scriptsize 76.2 & \scriptsize 99.4 & \scriptsize 93.8 & \scriptsize 83.0 & \scriptsize \textbf{89.8} & \scriptsize \textbf{77.7} & \scriptsize \textbf{59.7} & \scriptsize \textbf{43.1} & \scriptsize 94.9 & \scriptsize 78.4 & \scriptsize \textbf{72.1} & \scriptsize 79.8 & \scriptsize 59.3 & \scriptsize \textbf{86.0} & \scriptsize 95.7 & \scriptsize 72.3 & \scriptsize 65.2 & \scriptsize \textbf{96.1} & \scriptsize 67.5 & \scriptsize \textbf{76.9} & \scriptsize 99.6 & \scriptsize \textbf{85.8} & \scriptsize \textbf{80.4} \\
        
        \end{tabular}
\end{center}}\end{minipage}
}
\\

\subfloat[
\textbf{Impact of image transforms on zero-shot video classification performance.}
\label{tab: ablate_transforms_video_cls}
]{
\centering
\begin{minipage}{1\linewidth}{\begin{center}
\tablestyle{1.4pt}{1.2}
\begin{tabular}{l|c|cccc|c}
        method\ph{+} & \#Frames & \scriptsize UCF-101 & \scriptsize K-400  & \scriptsize K-600 & \scriptsize K-700 & \textbf{avg. acc.} \\
        \shline

        \scriptsize DFN5B-CLIP-H/14+* & 1 & \scriptsize 78.5 & \scriptsize 65.2 & \scriptsize 66.0 & \scriptsize 59.2 & \scriptsize 67.2 \\

        \scriptsize DFN5B-CLIP-H/14+$\dag$ & 1 & \scriptsize \textbf{79.2} & \scriptsize \textbf{66.7} & \scriptsize \textbf{67.0} & \scriptsize \textbf{60.7} & \scriptsize \textbf{68.4} \\

        \midrule
        \rgray
        \scriptsize \evaclipx * & 1 & \scriptsize \textbf{86.0} & \scriptsize 72.2 & \scriptsize 72.6 & \scriptsize 67.4 & \scriptsize 74.6 \\
        \rgray
        \scriptsize \evaclipx$\dag$ & 1 & \scriptsize 85.6 & \scriptsize \textbf{72.9} & \scriptsize \textbf{72.9} & \scriptsize \textbf{68.2} & \scriptsize \textbf{74.9} \\
        \midrule
        \rgray
        \scriptsize \evaclipx * & 8 & \scriptsize \textbf{88.2} & \scriptsize \textbf{79.3} & \scriptsize \textbf{79.2} & \scriptsize 72.0 & \scriptsize \textbf{79.7} \\
        \rgray
        \scriptsize \evaclipx$\dag$ & 8 & \scriptsize 87.9 & \scriptsize 79.2 & \scriptsize 79.1 & \scriptsize \textbf{72.1} & \scriptsize 79.6 \\
    \end{tabular}
\end{center}}\end{minipage}
}
\\
\subfloat[
\textbf{Impact of image transforms on zero-shot retrieval performance.}
\label{tab: ablate_transforms_retrieval}
]{
\centering
\begin{minipage}{1\linewidth}{\begin{center}
\tablestyle{1.4pt}{1.2}
    \begin{tabular}{l|ccc|ccc|ccc|ccc|c}
        & \multicolumn{6}{c|}{zero-shot \textbf{text} retrieval} & \multicolumn{6}{c|}{zero-shot \textbf{image} retrieval} \\
        & \multicolumn{3}{c|}{\scriptsize Flickr30K} & \multicolumn{3}{c|}{\scriptsize COCO} & \multicolumn{3}{c|}{\scriptsize Flickr30K} & \multicolumn{3}{c|}{\scriptsize COCO} \\
    
        method{\scriptsize{\ph{+}}} & \scriptsize R@1 & \scriptsize R@5 & \scriptsize R@10 & \scriptsize R@1 & \scriptsize R@5 & \scriptsize R@10 & \scriptsize R@1 & \scriptsize R@5 & \scriptsize R@10 & \scriptsize R@1 & \scriptsize R@5 & \scriptsize R@10 & MR \\
        \shline

        \scriptsize DFN5B-CLIP-H/14 +* & \scriptsize 92.3 & \scriptsize 99.1 & \scriptsize \textbf{99.7} & \scriptsize 70.6 & \scriptsize 89.6 & \scriptsize 94.4 & \scriptsize 80.7 & \scriptsize 95.5 & \scriptsize 97.7 & \scriptsize 54.1 & \scriptsize 78.0 & \scriptsize 85.4 & \scriptsize 86.4 \\
        
        \scriptsize DFN5B-CLIP-H/14 +$\dag$ & \scriptsize \textbf{93.6} & \scriptsize \textbf{99.3} & \scriptsize 99.6 & \scriptsize \textbf{71.8} & \scriptsize \textbf{90.4} & \scriptsize \textbf{94.9} & \scriptsize \textbf{82.1} & \scriptsize \textbf{96.0} & \scriptsize \textbf{97.9} & \scriptsize \textbf{55.6} & \scriptsize \textbf{79.2} & \scriptsize \textbf{86.3} & \scriptsize \textbf{87.2} \\

        \midrule

        \rgray
        \scriptsize \evaclipx* & \scriptsize 95.4 & \scriptsize 99.5 & \scriptsize 99.8 & \scriptsize 72.8 & \scriptsize 89.7 & \scriptsize 94.3 & \scriptsize 83.2 & \scriptsize 95.9 & \scriptsize 97.8 & \scriptsize 55.6 & \scriptsize 77.9 & \scriptsize 85.3 & \scriptsize 86.7 \\

        \rgray
        \scriptsize \evaclipx$\dag$ & \scriptsize \textbf{96.7} & \scriptsize \textbf{99.7} & \scriptsize \textbf{100.0} & \scriptsize \textbf{73.6} & \scriptsize \textbf{90.9} & \scriptsize \textbf{95.0} & \scriptsize \textbf{83.3} & \scriptsize \textbf{96.3} & \scriptsize \textbf{98.3} & \scriptsize \textbf{56.2} & \scriptsize \textbf{78.5} & \scriptsize \textbf{85.6} & \scriptsize \textbf{87.8} \\
        
    \end{tabular}
\end{center}}\end{minipage}
}
\caption{\textbf{Impact of image transformations on zero-shot evaluation.} $\dag$ denotes the direct resizing of images to a fixed size, while * indicates resizing images based on the shortest side and subsequently center cropping to achieve a fixed size.}
\label{tab: ablate_transforms}
\end{table*}

\section{Training Settings} 

We present detailed training settings of \evaclipeight and \evaclipx in~\cref{tab: clip cfg 1,tab: clip cfg 2}.

\section{Image Transformations for Evaluation} 

Two prevalent image transformations utilized in zero-shot evaluation are: 1) direct resizing of images to a fixed size, such as 224$\times$224, and 2) resizing images based on the shortest side, followed by center cropping to achieve a fixed size. In \tblref{tab: ablate_transforms}, our study systematically investigates the impact of these two image transformations in zero-shot evaluations. Notably, there exists a significant performance gap between the two transformations, observed particularly in zero-shot image classification on ObjectNet~\cite{objectnet} and VOC2007~\cite{pascal-voc-2007}, and zero-shot retrieval on Flickr30K~\cite{flickr30K} and COCO~\cite{lin2014coco}. \evaclipx shows robustness with almost the same average accuracy across different image transformations in zero-shot image/video classification.

For zero-shot image classification and video classification, we present results obtained by selecting the best-performing transformation between the two. In the case of zero-shot retrieval tasks, we specifically choose the transformation that involves direct resizing of images to a fixed size.

\end{document}